%% file: main.tex
\documentclass[runningheads]{llncs}

\usepackage[mobile]{eccv}

\usepackage{xcolor, colortbl}
\usepackage{multirow, multicol}
\usepackage{bbm}
\usepackage{makecell}

\usepackage{pifont}
\newcommand{\cmark}{\ding{51}}
\newcommand{\xmark}{\ding{55}}

\newcommand{\expnum}[2]{{#1}\mathrm{e}{-#2}}

\definecolor{Gray}{gray}{0.9}
\newcommand{\beginsupplement}{
    \setcounter{section}{0}
    \renewcommand{\thesection}{\Alph{section}}
    \renewcommand{\theHsection}{\Alph{section}}
    \setcounter{table}{0}
    \renewcommand{\thetable}{S\arabic{table}}
    \setcounter{figure}{0}
    \renewcommand{\thefigure}{S\arabic{figure}}}

\usepackage{eccvabbrv}

\usepackage{graphicx}
\usepackage{booktabs}

\usepackage[accsupp]{axessibility}  

\usepackage[colorlinks]{hyperref}

\usepackage{orcidlink}

\begin{document}

\title{Towards More Practical Group Activity Detection: A New Benchmark and Model} 

\titlerunning{Towards More Practical Group Activity Detection}

\author{Dongkeun Kim\inst{}\orcidlink{0000-0001-6093-126X}\qquad
Youngkil Song\inst{}\orcidlink{0009-0001-9565-2102}\qquad
Minsu Cho\inst{}\orcidlink{0000-0001-7030-1958}\qquad
Suha Kwak\inst{}\orcidlink{0000-0002-4567-9091}}

\authorrunning{Dongkeun Kim, Youngkil Song, Minsu Cho, and Suha Kwak}

\institute{Pohang University of Science and Technology (POSTECH), South Korea\\
\email{\{kdk1563,songyk,mscho,suha.kwak\}@postech.ac.kr}\\
\url{https://cvlab.postech.ac.kr/research/CAFE}}

\maketitle
\input{arxiv_sections/0_abstract}
\input{arxiv_sections/1_introduction}
\input{arxiv_sections/2_related_work}
\input{arxiv_sections/3_dataset}
\input{arxiv_sections/4_method}
\input{arxiv_sections/5_experiments}
\input{arxiv_sections/6_conclusion}

{\small
\noindent \textbf{Acknowledgement.} 
This work was supported by 
the NRF grant and 
the IITP grant 
funded by Ministry of Science and ICT, Korea
(RS-2019-II191906,          
IITP-2020-0-00842,          
NRF-2021R1A2C3012728,       
RS-2022-II220264).          
We thank Deeping Source for their help with data collection.
}

%
%
\bibliographystyle{splncs04}
\bibliography{cvlab_video}

\clearpage
\beginsupplement

\noindent
\textbf{\Large Appendix}
\vspace{1em}

\input{arxiv_supp_sections/0_introduction}
\input{arxiv_supp_sections/1_dataset}
\input{arxiv_supp_sections/2_experiments}

\end{document}

%% file: arxiv_sections/0_abstract.tex
\begin{abstract}
Group activity detection (GAD) is the task of identifying members of each group and classifying the activity of the group at the same time in a video.
While GAD has been studied recently, there is still much room for improvement in both dataset and methodology due to their limited capability to address practical GAD scenarios.
To resolve these issues, we first present a new dataset, dubbed Caf\'e.
Unlike existing datasets, Caf\'e is constructed primarily for GAD and presents more practical scenarios and metrics, as well as being large-scale and providing rich annotations.
Along with the dataset, we propose a new GAD model that deals with an unknown number of groups and latent group members efficiently and effectively.
We evaluated our model on three datasets including Caf\'e, where it outperformed previous work in terms of both accuracy and inference speed.
\keywords{Group activity detection \and Social group activity recognition}
\end{abstract}

%% file: arxiv_sections/1_introduction.tex
\section{Introduction}

Understanding group activities in videos plays a crucial role in numerous applications such as visual surveillance, social scene understanding, and sports analytics. 
The generic task of group activity understanding is complex and challenging since it involves identifying participants in an activity and perceiving their spatio-temporal relations as well as recognizing actions of individual actors.
Due to these difficulties, most existing work on group activity understanding has been limited to the task of categorizing an entire video clip into one of predefined activity classes~\cite{azar2019convolutional,wu2019learning,gavrilyuk2020actor,pramono2020empowering,yuan2021spatio,li2021groupformer,kim2022detector}, which is called the group activity recognition (GAR) in the literature.  
The common setting of GAR assumes that only a single group activity appears per clip and actors taking part of the activity are identified manually in advance. 
However, these assumptions do not hold in many real crowd videos, which often exhibit multiple groups 
that perform their own activities and \emph{outliers} who do not belong to any group. 
Moreover, it is impractical to manually identify the actors relevant to each group activity in advance.
Hence, although GAR has served as a representative group activity understanding task for a decade, its practical value is largely limited.

\begin{figure}[t]
\centering
\includegraphics[width=\textwidth]{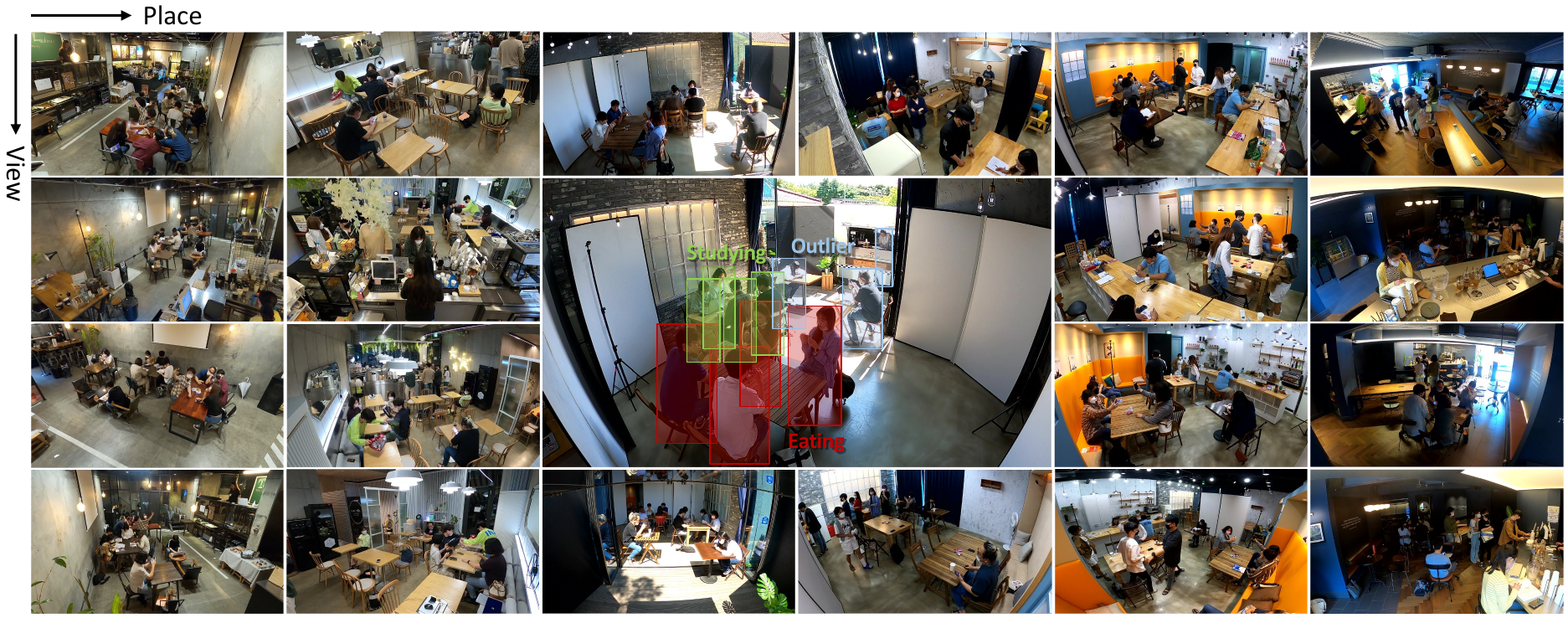}
\caption{
Examples of videos in Caf\'e. The videos were taken at six different places and four cameras with different viewpoints in each place.
}
\label{fig:fig1}
\end{figure}

As a step toward more realistic group activity understanding, group activity detection (GAD) has recently been studied~\cite{ehsanpour2020joint,ehsanpour2022jrdb,tamura2022hunting}.
GAD aims to localize multiple groups in a video clip and classify each of the localized groups into one of predefined group activity classes, where the group localization means identifying actors of each group.
Although a few prior work sheds light on this new and challenging task, there is still large room for improvement in both dataset and methodology due to their limited capability to address practical GAD scenarios.  
Existing datasets for GAD~\cite{ehsanpour2020joint,ehsanpour2022jrdb} are not constructed primarily for the task but are extensions of other datasets~\cite{choi2009they,martin2021jrdb} with additional group labels. 
Moreover, most of the groups in these datasets are singletons, which are individuals rather than groups.
Meanwhile, most GAD models rely on off-the-shelf clustering algorithms for group localization, which are not only too heavy in computation but also not optimized for the task.

To address the dataset issue, we present a new dataset for GAD, dubbed Caf\'e. 
Examples of videos in Caf\'e are presented in \cref{fig:fig1}.
The videos were taken at six different cafes where people tend to gather in groups, capturing realistic daily group activities. 
Each video exhibits multiple groups performing various activities, along with outliers, presenting more practical scenarios for GAD.
Caf\'e has several advantages over the existing GAD datasets.
First, it is significantly larger, providing 10K clips and 3.5M actor bounding box labels, as summarized in Table~\ref{table:tab1}.
Second, Caf\'e poses a greater challenge for group localization since it capture more densely populated scenes than the others; group localization on Caf\'e demands an accurate understanding of semantic relations between actors as well as their spatial proximity.
Finally, Caf\'e captures the same scene with up to four cameras from different viewpoints; these multi-view videos can be used to evaluate model's generalization on unseen views as well as unseen places.

\setlength{\tabcolsep}{4pt}
\begin{table}[t]
\caption{
Comparison between Caf\'e and other datasets for group activity understanding. 
`\# Clips' and `\# Boxes' represent the number of video clips and the number of annotated bounding boxes, respectively.} 
\label{table:tab1}
\centering
\small
\scalebox{0.9}{
\begin{tabular}{cccccccc}
\hline
Dataset     & \# Clips    & Resolution        & \# Boxes    & Source          & Multi-group    & Multi-view\\
\hline
CAD~\cite{choi2009they}                     
            & 2,511       & $720\times480$    & 0.1M        & Daily videos    & \xmark         & \xmark\\
Volleyball~\cite{ibrahim2016hierarchical}
            & 4,830       & $1280\times720$   & 1.2M        & Sports videos   & \xmark         & \xmark\\
NBA~\cite{yan2020social}
            & 9,172       & $1280\times720$   & -           & Sports videos   & \xmark         & \xmark\\
\hline
PLPS~\cite{qing2021public}
            & 71          & $1920\times1080$  & 0.2M        & Daily videos    & \cmark         & \xmark\\
Social-CAD~\cite{ehsanpour2020joint}
            & 2,511       & $720\times480$    & 0.1M        & Daily videos    & \cmark         & \xmark\\
JRDB-Act~\cite{ehsanpour2022jrdb}
            & 3,625       & $3760\times480$   & 2.5M        & Daily videos    & \cmark         & \xmark\\
\rowcolor{Gray}
Caf\'e                                      
            & 10,297      & $1920\times1080$  & 3.5M        & Daily videos    & \cmark         & \cmark\\
\hline
\end{tabular}}
\end{table}

In addition to the new dataset, we also propose a new model architecture for end-to-end GAD. 
Our model builds embedding vectors of group candidates and individual actors through the attention mechanism of Transformer~\cite{vaswani2017attention,dosovitskiy2021an}.
Unlike GAR approaches using Transformer, which aggregate actor features to form a single group representation while capturing spatio-temporal relationship between actors, our model divides actors into multiple groups, each with its own group representation.
Embedding vectors of an actor and a group are learned to be close to each other if the actor is a member of the group so that group localization is done by matching the actor and group embeddings.
To deal with an unknown number of groups, we employ learnable group tokens whose number is supposed to be larger than the possible maximum number of groups in a video clip; 
the tokens are then transformed into group embeddings by Transformer, attending to actor embeddings.
Each group embedding is also used as input to an activity classifier that determines its activity class.
This mechanism allows to discover groups accurately without off-the-shelf clustering algorithms unlike most of previous work, leading to substantially faster inference.

We evaluated our model on three datasets, Caf\'e, Social-CAD~\cite{ehsanpour2020joint}, and JRDB-Act~\cite{ehsanpour2022jrdb}, where it outperformed previous work in terms of both accuracy and inference speed.
In summary, our contribution is three-fold as follows: 
\begin{itemize}
    \item[$\bullet$] We introduce Caf\'e, a new challenging dataset for GAD. Thanks to its large-scale, rich annotations, densely populated scenes, and multi-view characteristics, it can serve as a practical benchmark for GAD.
    \item[$\bullet$] We present a novel GAD model based on Transformer that localizes groups based on the similarity between group embeddings and actor embeddings. 
    Our model efficiently deals with an unknown number of groups and latent group members without off-the-shelf clustering algorithms.
    \item[$\bullet$] Our model outperformed previous work on Caf\'e and two other GAD benchmarks in terms of both GAD accuracy and inference speed. 
\end{itemize}

%% file: arxiv_sections/2_related_work.tex
\section{Related Work}
\subsection{Group Activity Recognition}
Group activity recognition (GAR) has been extensively studied as a representative group activity understanding task. 
With the advent of deep learning, recurrent neural networks have substantially improved GAR performance~\cite{deng2016structure,ibrahim2016hierarchical,bagautdinov2017social,li2017sbgar,shu2017cern,wang2017recurrent,qi2018stagnet,ibrahim2018hierarchical,yan2018participation}.
In particular, hierarchical long short-term memory networks~\cite{wang2017recurrent,ibrahim2018hierarchical,yan2018participation} have been used to model the dynamics of individual actors and aggregate actors to infer the dynamics of a group.

A recent trend in GAR is modeling spatio-temporal relations between actors.
To this end, graph neural networks (GNN)~\cite{wu2019learning,ehsanpour2020joint,hu2020progressive,yan2020higcin,yuan2021spatio}, have been placed on top of a convolutional neural network (CNN).
Popular examples of such modules include graph convolutional networks~\cite{wu2019learning}, graph attention networks~\cite{ehsanpour2020joint}, dynamic relation graphs~\cite{yuan2021spatio}, and causality graphs~\cite{xie2023actor,zhang2024bi}.
To employ global spatio-temporal dynamic relations between actors and contexts,
Transformers~\cite{vaswani2017attention,dosovitskiy2021an} have been adopted for GAR and shown significant performance improvement~\cite{gavrilyuk2020actor,li2021groupformer,kim2022detector,pramono2020empowering,yuan2021learning,li2022learning,zhou2022composer,han2022dual}.
They utilize the attention mechanism to employ spatio-temporal actor relations~\cite{gavrilyuk2020actor,li2022learning,han2022dual}, relational contexts with conditional random fields~\cite{pramono2020empowering}, actor-specific scene context~\cite{yuan2021learning}, intra- and inter-group contexts~\cite{li2021groupformer}, and partial contexts of a group activity~\cite{kim2022detector}.
Although these methods have demonstrated outstanding performance, since GAR assumes that only one group is present in each video, their applicability in real-world scenarios is substantially limited.

\subsection{Group Activity Detection}
Group activity detection (GAD), which is closely related to social group activity recognition~\cite{ehsanpour2020joint,ehsanpour2022jrdb,tamura2022hunting} and panoramic activity recognition (PAR)~\cite{han2022panoramic}, has recently been studied to address the limitation of GAR.
GNNs~\cite{ehsanpour2020joint,ehsanpour2022jrdb,han2022panoramic} have been utilized to model relations between actors and to divide them into multiple groups by applying graph spectral clustering~\cite{ng2001spectral,zelnik2004self}.
However, they require off-the-shelf clustering algorithms, which are not optimized for the task and resulting slow inference speed.
Meanwhile, HGC~\cite{tamura2022hunting}, which is most relevant to our work, adopts Deformable DETR~\cite{zhu2020deformable} for localization and matches a group and its potential members in 2D coordinate space.
Unlike HGC, our model conducts such a matching in an embedding space to exploit semantic clues more explicitly, achieving better performance.

Along with these models, several datasets have been introduced.
Social-CAD~\cite{ehsanpour2020joint} extends CAD~\cite{choi2009they} by adding sub-group labels. 
JRDB-Act~\cite{ehsanpour2022jrdb} and JRDB-PAR~\cite{han2022panoramic} extend annotations of JRDB~\cite{martin2021jrdb}, a multi-person dataset captured by a mobile robot with panoramic views.
On these datasets, actors are divided into multiple groups, and the activity of each group is determined by majority voting of individual actions.
However, most of the groups in these datasets are composed of a single actor, which is an individual who does not interact with other actors.
Unlike these datasets, Caf\'e is constructed primarily for GAD.
Also, in Caf\'e, people annotated as a group perform an activity together, and singleton groups are annotated as outliers.

%% file: arxiv_sections/3_dataset.tex
\section{Caf\'e Dataset}

Caf\'e is a multi-person video dataset that aims to introduce a new challenging benchmark for GAD.
The dataset contains more than 4 hours of videos taken at six different cafes by four cameras with different viewpoints, and provides rich annotations including 3.5M bounding boxes of humans, their track IDs, group configurations, and group activity labels.
In an untrimmed video, an actor can engage in varying group activities over time, which makes the task challenging and comparisons with existing methods infeasible.
Thus, each of the videos is segmented into 6-second clips.
In each clip, each actor is a member of a group that performs one of six different group activities or is an \textit{outlier} who does not belong to any group (\ie, a singleton group).
Also, outliers are often located overly close to groups as shown in \cref{fig:fig1}.
Thus, for group localization in Caf\'e, it is required to grasp the properties of individual actors and their semantic relations as well as their spatial proximity.

\subsection{Dataset Annotation and Splitting}
\label{sec:dataset_annotation}
Human annotators selected the key frame that clearly exhibited group activities in each video clip. Then, they annotated actor bounding boxes, group configurations, and group activity labels in the frame.
Next, a multi-object tracker~\cite{zhang2021bytetrack} was applied to extend the actor box labels from the key frame into tracklets across the frames of the clip.
To improve the quality of estimated tracklets, the tracker utilized a person detector~\cite{ge2021yolox} pretrained on public datasets for person detection and tracking such as CrowdHuman~\cite{shao2018crowdhuman}, MOT17~\cite{dendorfer2021motchallenge}, 
City Person~\cite{zhang2017citypersons}, and ETHZ~\cite{ess2008mobile}, which was further finetuned using the key frames of Caf\'e.
Finally, the annotators manually fixed incorrect tracking IDs and box coordinates. 

To examine both place and viewpoint generalization of tested models, we split the dataset in two different ways: \textit{split by place} and \textit{split by view}. 
The \textit{split by view} setting demonstrates the multi-view characteristics of Caf\'e by evaluating the model on unseen views, a challenge absent in existing GAD benchmarks.
Details of each dataset split is provided in the appendix (\cref{sec:a.2}).

\subsection{Dataset Statistics}
\label{sec:dataset_statistics}

\begin{figure}[t!]
    \centering    
    \subfloat[\centering \label{fig:class_dist}]{{\includegraphics[width=0.48\linewidth]{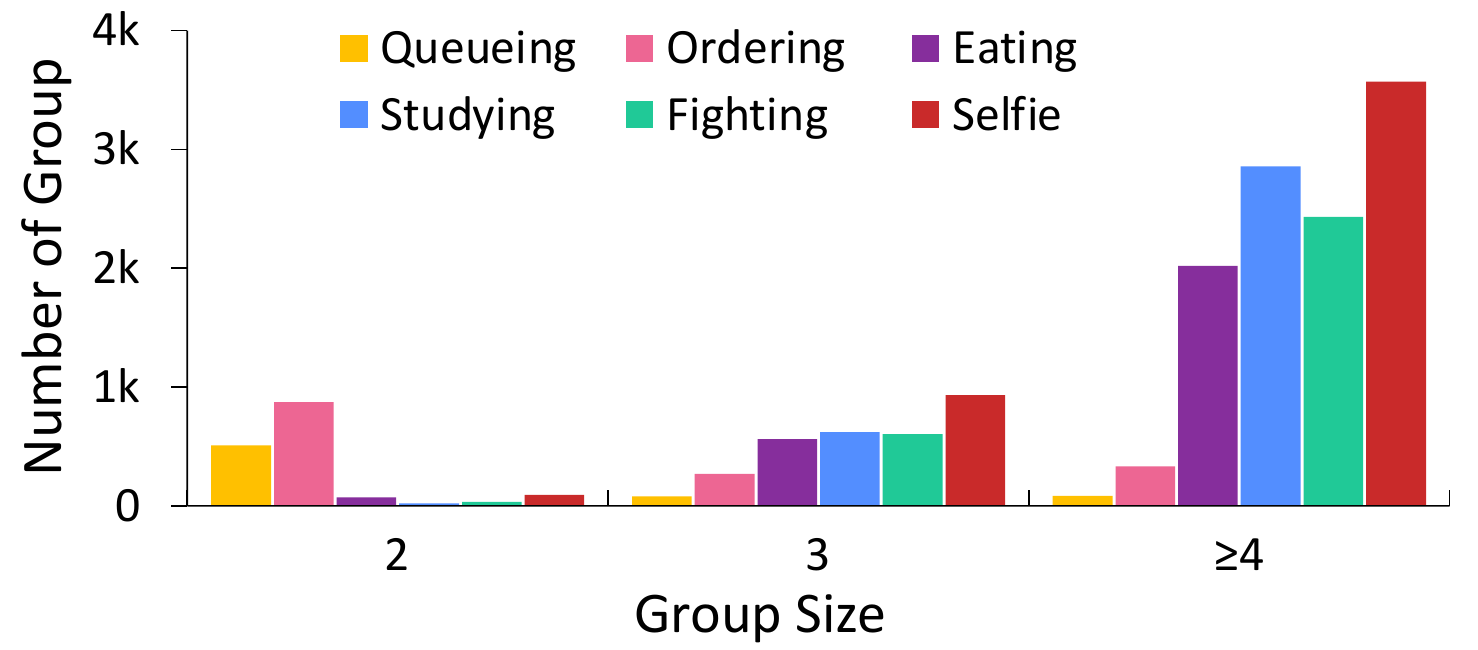}}}    
    \hfill
    \subfloat[\centering \label{fig:num_actors}]{{\includegraphics[width=0.48\linewidth]{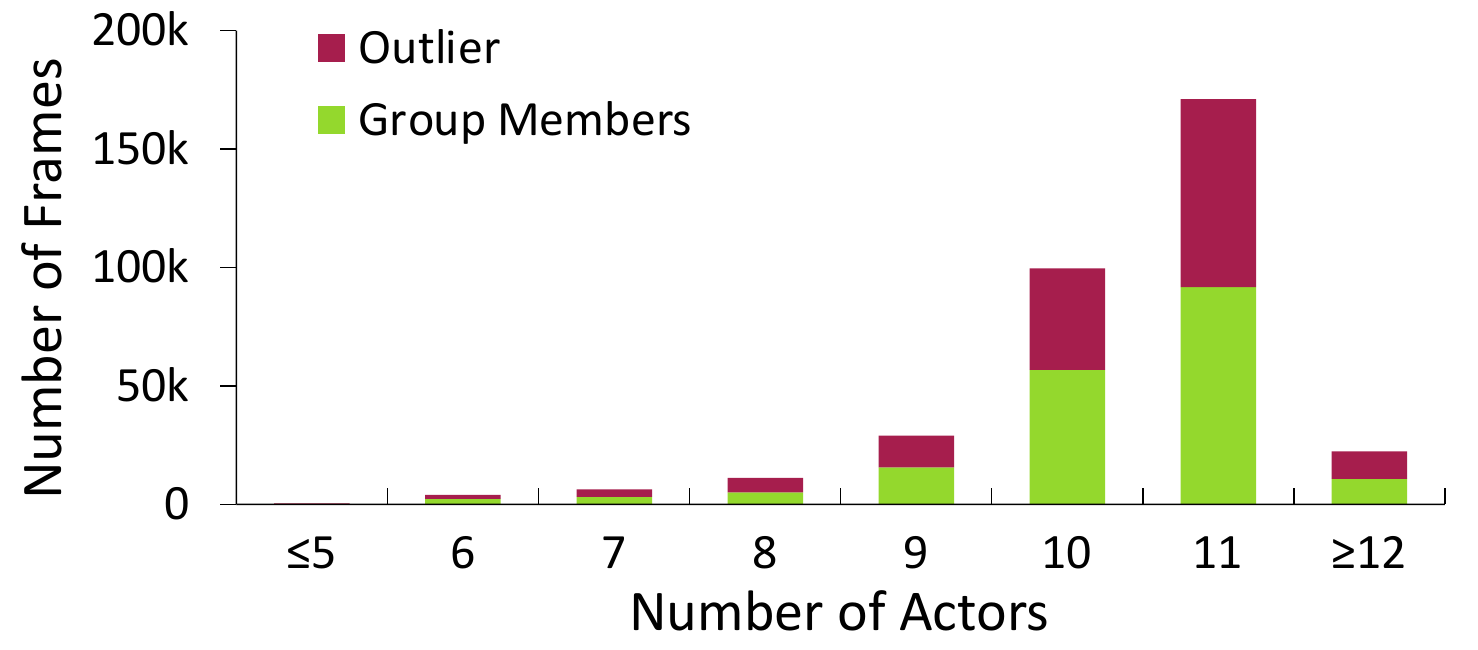}}}    
    \hfill
    \caption{A summary statistics of Caf\'e.
    (a) Group population versus group size per activity class.
    (b) Distribution of the number of actors in each video frame.
    }
    \label{fig:dataset_stat}
\end{figure}

Important statistics that characterize Caf\'e are summarized in \cref{fig:dataset_stat}.
\cref{fig:class_dist} shows group population versus group size (\ie, the number of group members) for each activity class. 
The class distribution of Caf\'e is imbalanced: The least frequent group activity \textit{Queueing} appears about seven times fewer than the most frequent group activity \textit{Taking Selfie}.
Such an imbalance is natural in the real world, and may deteriorate activity classification accuracy. 

As shown in \cref{fig:num_actors}, the number of actors in each video clip varies from 3 to 14, and most clips contain 10 or 11 actors. 
We thus argue that videos in Caf\'e well simulate real crowd scenes. 
Also, about half of the actors are outliers in each clip, which suggests that, on Caf\'e, group localization is more challenging.

\begin{figure}[t!]
    \centering  
    \subfloat[\centering \label{fig:group_size}]{{\includegraphics[width=0.48\linewidth]{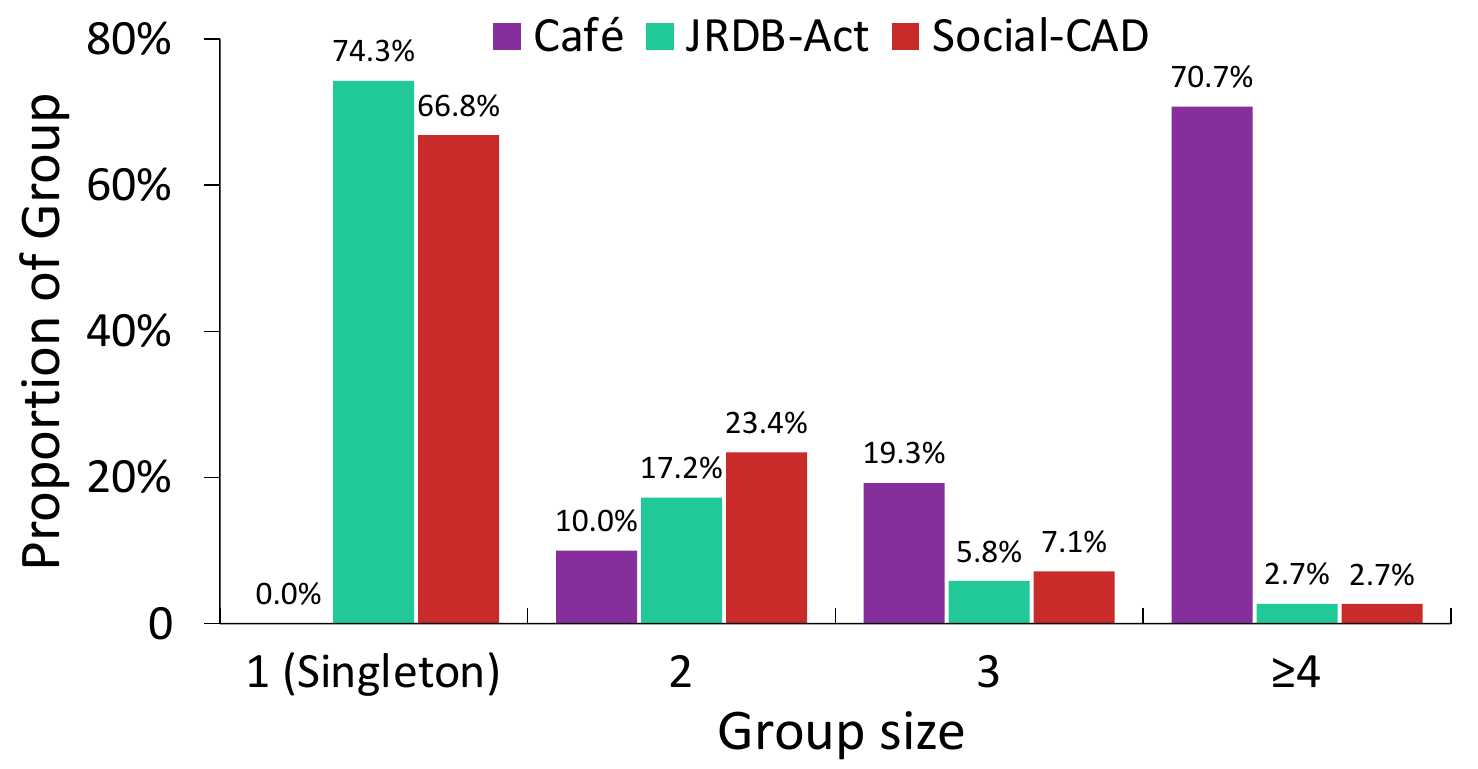}}}
    \hfil
    \subfloat[\centering \label{fig:aspect_ratio}]{{\includegraphics[width=0.48\linewidth]{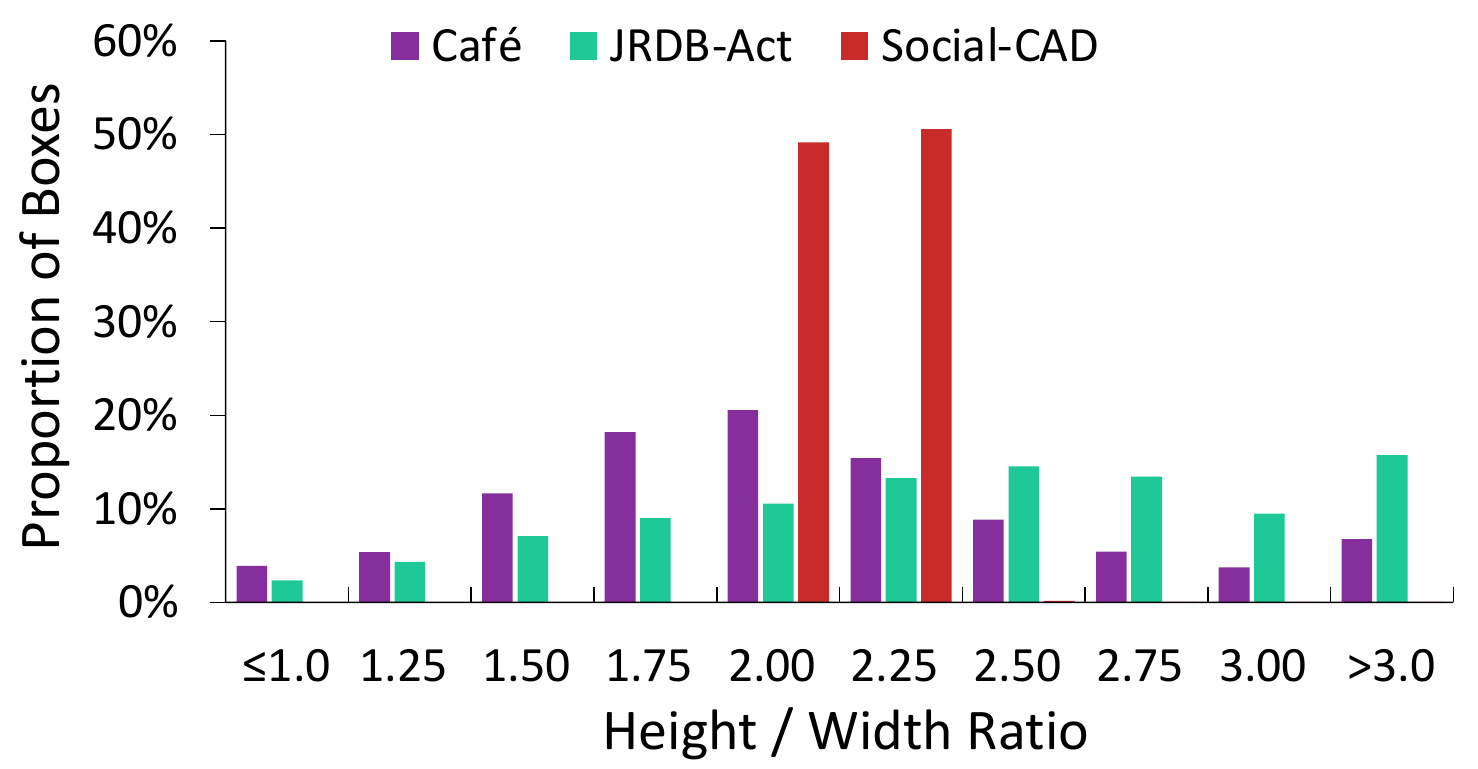}}}
    \hfil
    \subfloat[\centering \label{fig:population_density}]{{\includegraphics[width=0.48\linewidth]{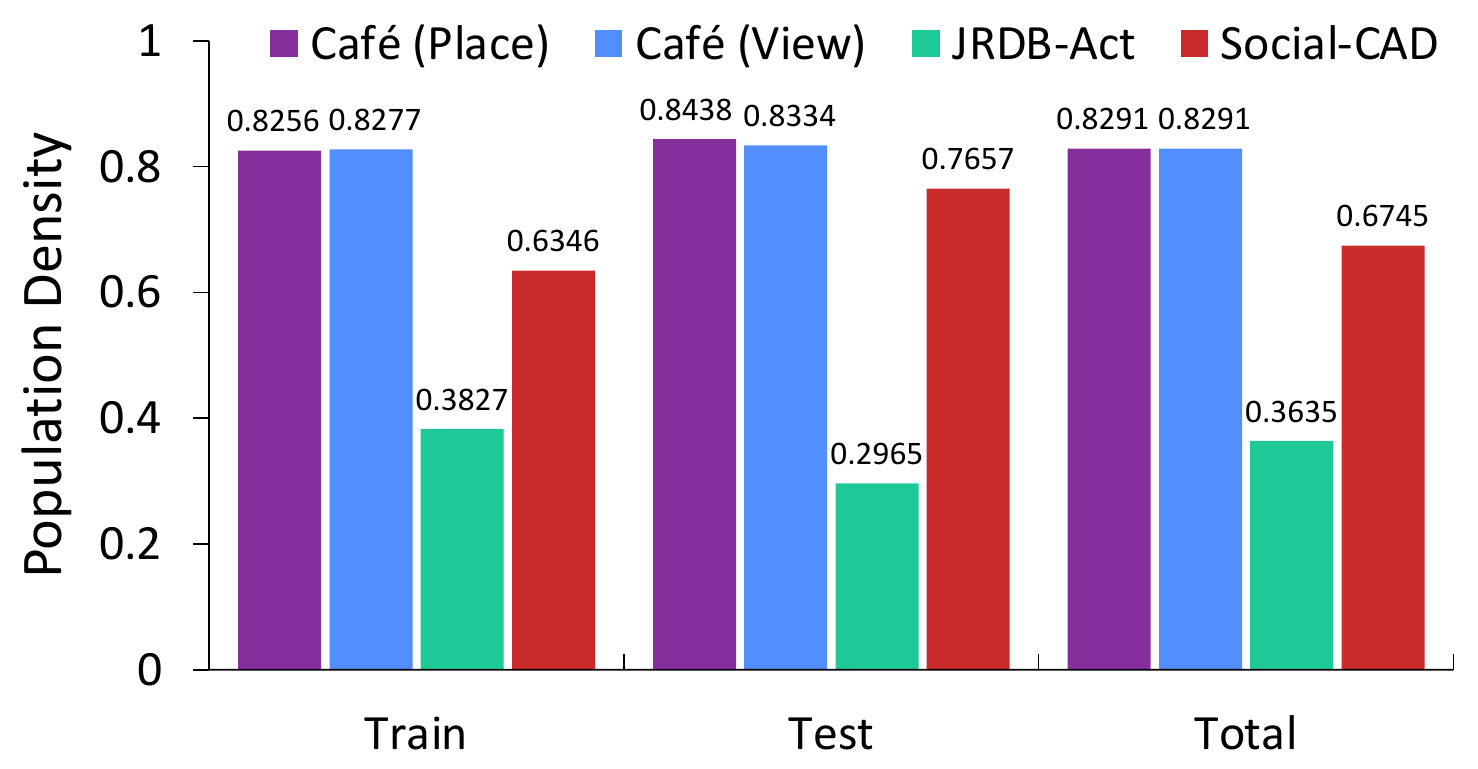}}}
    \hfil
    \subfloat[\centering \label{fig:inter_distance}]{{\includegraphics[width=0.48\linewidth]{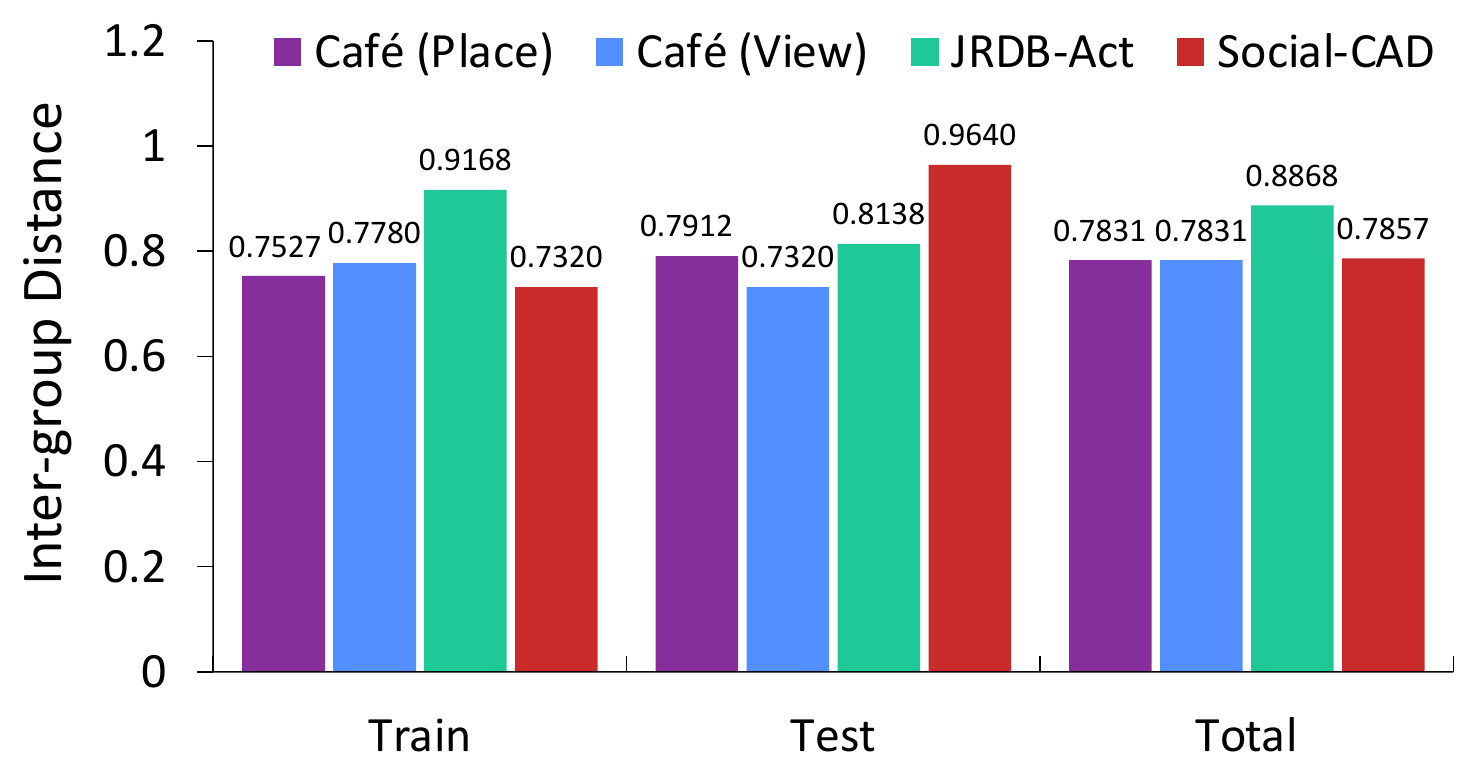}}}
    \hfil
    \caption{Comparison between Caf\'e and existing GAD datasets in terms of
    (a) group size,
    (b) aspect ratios of actor boxes, 
    (c) population density, and
    (d) inter-group distance.
    }
    \label{fig:dataset_comparison}
\end{figure}

\subsection{Comparison with Existing GAD Datasets}
\label{sec:dataset_comparison}

To show the unique challenges and practical aspects of Caf\'e, we compare Caf\'e with the existing GAD datasets, Social-CAD~\cite{ehsanpour2020joint} and JRDB-Act~\cite{ehsanpour2022jrdb}.
\cref{fig:group_size} shows that most groups of existing datasets comprise only a single actor, which are not actually groups but individuals.
On the other hand, all groups in Caf\'e have at least two actors, and mostly contain more than or equal to four actors.

\cref{fig:aspect_ratio} illustrates the aspect ratio distribution of actor bounding boxes.
In Social-CAD, actors are predominantly pedestrians moving or standing, resulting in nearly all aspect ratios being around $1:2$. 
In contrast, Caf\'e and JRDB-Act present diverse group activities, resulting in significant pose variation and diverse aspect ratios. 
Particularly, activities like \textit{Fighting} and \textit{Taking Selfie} in Caf\'e necessitates capturing fine-grained pose information, making it a more challenging.

We also compare the datasets in terms of population density, which we define as the ratio between the union area of actors participating in group activities and the area of their convex hull.
As shown in \cref{fig:population_density}, Caf\'e exhibits a higher population density compared to the others in both the \textit{split by view} and \textit{split by place}, making it more challenging for detecting group activities.

Finally, \cref{fig:inter_distance} compares the datasets in the inter-group distance, which calculated by computing the distance between each group and its nearest group or outlier and taking the average of such distances;
the exact formulation of the inter-group distance can be found in the appendix (\cref{sec:inter_distance}).
A lower inter-group distance indicates that groups are harder to be localized only by spatial proximity, making the benchmark more challenging
than the other two datasets.

\subsection{Evaluation Metrics}
\label{sec:metric}
A proper evaluation metric for GAD should consider following two aspects of predictions: (1) group localization, \ie, identification of members per group, and (2) activity classification per group. 
While a few evaluation metrics such as social accuracy, social mAP, and G-Act mAP were already proposed in previous work~\cite{ehsanpour2020joint,ehsanpour2022jrdb}, they evaluate group localization based on individual actors rather than groups, which makes them less strict criterion for evaluating group localization quality. 

Hence, we propose new evaluation metrics for GAD: \emph{Group mAP} and \emph{Outlier mIoU}. 
Group mAP is a modification of mAP that has been widely used as the standard performance metric for object detection.
On the other hand, Outlier mIoU evaluates how much correctly a model identifies outliers of input video.

\noindent
\textbf{Group mAP.}
Before introducing the definition of Group mAP, we first define Group IoU~\cite{choi2014discovering}, analogous to IoU used in computation of mAP for object detection.
Group IoU measures group localization accuracy by comparing a ground-truth group and a predicted group as follows:
\begin{equation}
    \text{Group IoU} (G, \hat{G}) = 
    \frac{|G \cap \hat{G}|}{|G \cup \hat{G}|},
    \label{eqd:giou}
\end{equation}
where $G$ is a ground-truth group and $\hat{G}$ is a predicted group; both groups are sets of actors.
Group IoU is 1 if all members of $\hat{G}$ are exactly the same with those of $G$ and 0 if no member co-occurs between them. 
$\hat{G}$ is considered as a correctly localized group if there exists a ground-truth group $G$ that holds $\text{Group IoU} (G, \hat{G}) \geq \theta$, where $\theta$ is a predefined threshold.
Note that we use two thresholds, $\theta=1.0$ and $\theta=0.5$, for evaluation.
Group mAP is then defined by using Group IoU as a localization criterion along with activity classification scores.
To be specific, we utilize the classification score of the ground-truth activity class as the detection confidence score of the predicted group, and calculate average precision (AP) score per activity class through all-point interpolation~\cite{everingham2015pascal}.  
Finally, AP scores of all classes are averaged to produce Group mAP.

\noindent
\textbf{Outlier mIoU.}
It is important for GAD in the real world videos to distinguish groups and outliers (\ie, singletons).
We thus propose Outlier mIoU to evaluate outlier detection.
Similar to Group IoU, its format definition is given by
\begin{align}
   \text{Outlier mIoU} = \frac{1}{|V|} \sum_{v\in V} \frac{|O_v \cap \hat{O}_v|}{|O_v \cup \hat{O}_v|},
\end{align}
where $|V|$ is the set of video clips for evaluation, $O_v$ is the set of ground-truth outliers in clip $v$, and $\hat{O}_v$ is the set of predicted outliers in clip $v$. 

%% file: arxiv_sections/4_method.tex
\section{Proposed Model for GAD}
\begin{figure}[t!]
\centering
\includegraphics[width=1.0\linewidth]{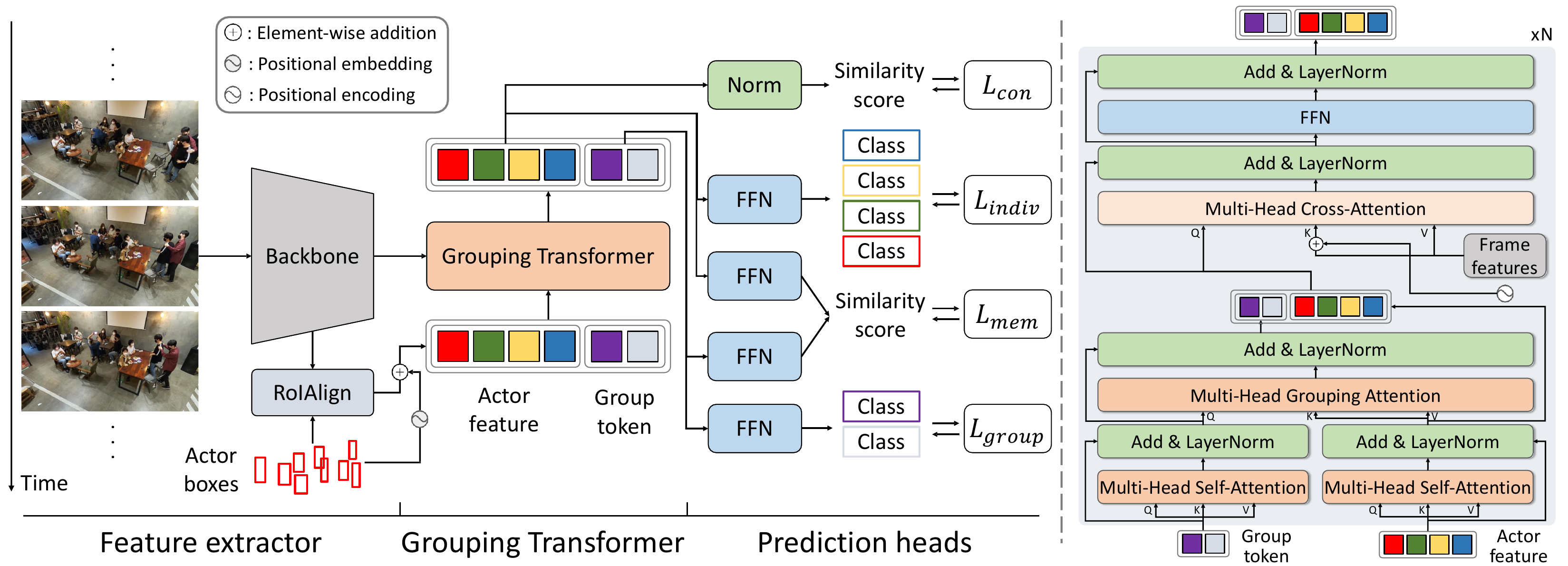}
\caption{(\textit{Left}) Overall architecture of our model. 
(\textit{Right}) Detailed architecture of the Grouping Transformer. }
\label{fig:architecture}
\end{figure}

The purpose of GAD is to identify members of each group (\ie, group localization) and classify the activity conducted by each group simultaneously.
The task is challenging since both the number of groups and their members are unknown.
We present a new model based on Transformer~\cite{vaswani2017attention,dosovitskiy2021an} to deal with these difficulties; its overall architecture is illustrated in \cref{fig:architecture}.

The key idea at the heart of our model is that embedding vectors of an actor and a group should be close if the actor is a member of the group. 
To compute embedding vectors of groups and individual actors, we adopt attention mechanism of Transformer.
To deal with a varying number of groups in each video clip, our model defines and utilizes learnable group tokens, whose number is supposed to be larger than the possible maximum number of groups in a clip.
The group tokens along with actor features obtained by RoIAlign~\cite{mask_rcnn} are fed to a Transformer called Grouping Transformer to become the embedding vectors.

\subsection{Model Architecture}
Our model consists of three parts: feature extractor, Grouping Transformer, and prediction heads. 

\noindent
\textbf{Feature extractor.}
As in recent GAR models~\cite{wu2019learning,gavrilyuk2020actor,ehsanpour2020joint,yuan2021learning,li2021groupformer,ehsanpour2022jrdb}, our model extracts frame-level features using a CNN backbone, and extracts actor features from the frame features by RoIAlign given actor bounding boxes. 
To be specific, we adopt an ImageNet~\cite{5206848} pretrained ResNet-18~\cite{resnet} for the feature extraction, and actor features extracted by RoIAlign are of $5\times5$ size. 
Additionally, to incorporate spatial cues when identifying group members, learnable positional embeddings of actor box coordinates are added to their associated actor features.

\noindent
\textbf{Grouping Transformer.}
Grouping Transformer takes learnable group tokens, actor features, and frame features as input, and produces embedding vectors of group candidates and actors in a frame-wise manner. 
As illustrated in the right-hand side of \cref{fig:architecture}, it comprises three types of multi-head attention layers:
(1) multi-head self-attention layers that capture relations between actors and those between groups separately,
(2) multi-head grouping attention layers where group tokens as queries attend to actor features serving as keys and values, and
(3) multi-head cross-attention layers where actor features and group tokens draw attentions on frame features to capture contextual information.
The core of the Grouping Transformer lies in the grouping attention layer. 
Each group token produces group representation by attending to actor features potentially belonging to its group, based on the similarity in the embedding space.
In addition, to exploit spatial cues, we apply a distance mask to the multi-head self-attention layers for actor features: Following ARG~\cite{wu2019learning}, a pair of actors whose distance is greater than a threshold $\mu$ do not attend to each other.

\noindent
\textbf{Prediction heads.}
Two types of prediction heads in the form of feed-forward networks (FFNs) are attached to individual outputs of the Grouping Transformer, actor embeddings and group embeddings. 
The first prediction heads are for group activity classification, and the second prediction heads further project the actor/group embeddings so that the results are used for identifying group members: 
An actor embedding and a group embedding projected separately are dot-producted to compute their semantic affinity, which is used as the membership score of the actor for the group. 
At inference, each actor is assigned to the group with the highest membership score among all predicted groups.

\subsection{Training Objectives}
\label{sec:4.2}
\noindent
\textbf{Group matching loss.}
Motivated by DETR~\cite{carion2020end}, 
we first establish the optimal bipartite matching between ground-truth groups and predicted groups using Hungarian algorithm~\cite{kuhn1955hungarian}. 
Since our model produces $K$ predicted groups, where $K$ is the number of group tokens and is supposed to be larger than the number of ground-truth groups, 
we add empty groups with no activity class, denoted by $\varnothing$, to the set of ground-truth groups so that the number of ground-truth groups becomes $K$ and they are matched with the predicted groups in a bipartite manner accordingly.
Then, among all possible permutations of $K$ predicted groups, denoted by $\mathfrak{S}_K$, Hungarian algorithm finds the permutation with the lowest total matching cost:
\begin{equation}
    \hat{\sigma} = \underset{\sigma \in \mathfrak{S}_K}{\arg\min} \sum_{i}^{K} C_{i, \sigma(i)}. \label{eq:hungarian}
\end{equation}
$C_{i, \sigma(i)}$ in Eq.~\eqref{eq:hungarian} is the matching cost of the ground-truth group $i$ and the predicted group $\sigma(i)$ and is given by
\begin{equation}
C_{i, \sigma(i)} =
-\mathbbm{1}_{\{y_i \neq \varnothing\}}\hat{p}_{\sigma(i)}(y_i) 
+\mathbbm{1}_{\{y_i \neq \varnothing\}}\| \textbf{m}_i - \hat{\textbf{m}}_{\sigma(i)} \|_{2}, \label{eqn:ind_matching_cost}
\end{equation}
where $y_i$ is the activity class label of the ground-truth group $i$ and $\hat{p}_{\sigma(i)}(y_i)$ is the predicted class probability for $y_i$.
Also, $\textbf{m}_i=[m_i^1, \ldots, m_i^N]^\mathsf{T}$ indicates ground-truth membership relations between actors and group $i$, and $\hat{\textbf{m}}_{\sigma(i)}=[\hat{m}_{\sigma(i)}^1,\ldots,\hat{m}_{\sigma(i)}^N]^\mathsf{T}$ is a collection of predicted membership scores of actors for predicted group $\sigma(i)$, where $N$ is the number of actors in the input clip;
each dimension of the two vectors is computed by
\begin{align}
    m_{i}^{j} & = 
    \begin{cases}
        1, & \text{if actor $j$ is a member of group $i$},\\
        0, & \text{otherwise},\\
    \end{cases} \label{eqn:eq3} \\
    \hat{m}_{\sigma(i)}^{j} & = {\psi_{j}}^\mathsf{T} \phi_{\sigma(i)}, 
\end{align}
where $\psi_j$ is the output of the second prediction head for actor $j$ and $\phi_{\sigma(i)}$ is the output of the second prediction head for predicted group $\sigma(i)$.
The group activity classification loss $L_\textrm{group}$ and the membership loss $L_\textrm{mem}$ are calculated for all matched pairs. 
To be specific, we adopt the standard cross-entropy loss for $L_\text{group} = L_\textrm{group}(i,\sigma(i))$:
\begin{equation}
    L_\text{group} = 
    - \log \frac{\exp (\hat{p}_{\sigma(i)} (y_i))}{\sum_{c=1}^{C} \exp (\hat{p}_{\sigma(i)} (c))},
    \label{eqn:group_loss}
\end{equation}
where $C$ is the number of group activity classes,
and the binary cross-entropy loss for $L_\text{mem} = L_\textrm{mem}(i,\sigma(i))$:
\begin{equation}
    L_\text{mem} = 
    - \frac{1}{N} \sum_{j=1}^{N} \big(m_i^j \cdot \log \hat{m}_{\sigma(i)}^j + (1 - m_i^j) \cdot \log (1 - \hat{m}_{\sigma(i)}^j)\big).
    \label{eqn:mem_loss}
\end{equation}

\noindent
\textbf{Group consistency loss.}
It has been known that supervisory signals given by the bipartite matching of Hungarian algorithm may fluctuate and thus lead to slow convergence~\cite{li2022dn}.
To alleviate this issue, we additionally introduce a group consistency loss, which is a modification of InfoNCE~\cite{oord2018representation} and enhances the quality of group localization while bypassing the bipartite matching. 
The loss is formulated by
\begin{equation}
L_\text{con} = - \sum_{g_i} \sum_{j \in g_i} \log \frac{\sum_{k \in g_{i}, k \neq j}\exp\big(\cos(f_j, f_k) / \tau\big)}{\sum_{k \neq j}\exp\big(\cos(f_j, f_k) / \tau\big)},
\label{eqn:eq1}
\end{equation}
where $g_i$ means the $i$-th ground-truth group, $\tau$ is the temperature, $f_j$ stands for $j$-th actor embeddings and $\cos$ indicates the cosine similarity function. 
This loss provides a consistent group supervision to actors that belong to the same group.

\noindent
\textbf{Individual action classification loss.}
We adopt a standard cross-entropy loss for individual action loss $L_{ind}$. 
The individual action class of an actor who belongs to a group is regarded as the group activity class of the group, and the action class of an outlier is no activity class, denoted by $\varnothing$. 

\noindent
\textbf{Total loss.}
Our model is trained with four losses simultaneously in an end-to-end manner. 
Specifically, the total training objective of our proposed model is a linear combination of the four losses as follows:
\begin{equation}
L = L_\text{ind} + \sum_{i}L_\text{group} + \lambda_{m} \sum_{i}L_\text{mem} + \lambda_{c} L_\text{con}.
\label{eqn:eq2}
\end{equation}

%% file: arxiv_sections/5_experiments.tex
\section{Experiments}

\subsection{Implementation Details}
\noindent
\textbf{Hyperparameters.}
We use an ImageNet pretrained ResNet-18 as a backbone network. 
Ground-truth actor tracklets are used to extract actor features with 256 channels by applying RoIAlign with crop size $5\times5$. 
For the Grouping Transformer, we stack 6 Transformer layers with 4 attention heads for Caf\'e and JRDB-Act, 3 Transformer layers with 8 attention heads for Social-CAD.
The number of group tokens $K$ is set to $12$ for Caf\'e and JRDB-Act, $10$ for Social-CAD.

\noindent
\textbf{Training.}
We sample $T$ frames using the segment-based sampling~\cite{wang2016temporal}, where $T$ is 5, 1, and 2 for Caf\'e, Social-CAD, and JRDB-Act, respectively.
We train our model with Adam optimizer~\cite{Adamsolver} with $\beta_{1}=0.9$, $\beta_{2}=0.999$, and $\epsilon=\expnum{1}{8}$ for 30 epochs. 
Learning rate is initially set to $\expnum{1}{5}$ with linear warmup to $\expnum{1}{4}$ for 5 epochs, and linearly decayed for remaining epochs. 
Mini-batch size is set to 16. 
Loss coefficients are set to $\lambda_{m}=5.0$, and $\lambda_{c}=2.0$. 
The temperature $\tau$ is set to $0.2$ for the group consistency loss.

\begin{table}[t]
\caption{
Comparison with the previous work on Caf\'e. 
All methods are built on the same ResNet-18 backbone.
The second column means the number of tokens for Transformer-based models and the number of clusters for clustering-based models. `\# G' and `\# O' indicates the true number of groups and that of outliers in each video clip, respectively. The third column is the wall-clock inference time for a single video clip measured on a Titan XP GPU. 
The subscripts of Group mAP mean Group IoU thresholds ($\theta$ in \cref{sec:metric}).
We mark the best and the second-best performance in \textbf{bold} and \underline{underline}, respectively.}
\label{table:gad_performance}
\centering
\scalebox{0.86}{
\begin{tabular}{lcccccccc}
    \hline
    \multirow{3}{*}{Method}  & \multirow{3}{*}{\makecell[c]{\# Token \\ (\# Cluster)}}  & \multirow{3}{*}{\makecell[c]{Inference \\ time (s)}}
    & \multicolumn{3}{c}{Split by view} & \multicolumn{3}{c}{Split by place}\\
    \cline{4-9}    
    & & & \makecell[c]{Group \\ mAP$_{1.0}$}    & \makecell[c]{Group \\ mAP$_{0.5}$}    & \makecell[c]{Outlier \\ mIoU}     
    & \makecell[c]{Group \\ mAP$_{1.0}$}    & \makecell[c]{Group \\ mAP$_{0.5}$}    & \makecell[c]{Outlier \\ mIoU}     \\
    \hline
    \multirow{4}{*}{ARG~\cite{wu2019learning}}
    & 4             & 0.22      & 11.03     & 34.50     & 56.61     & 6.87      & 28.44     & 46.72 \\
    & 5             & 0.26      & 5.46      & 30.34     & 58.89     & 5.79      & 24.25     & 49.25 \\
    & 6             & 0.28      & 1.27      & 27.69     & 60.41     & 2.59      & 22.33     & 51.00 \\
    & \#G+\#O   & 0.30      & 2.64      & 28.98     & 58.21     & 2.29      & 22.33     & 50.01 \\
    \hline
    \multirow{4}{*}{Joint~\cite{ehsanpour2020joint}}
    & 4             & 0.23      & 13.86     & 34.68     & 53.67     & 6.69      & 27.76     & 49.50 \\
    & 5             & 0.25      & 14.05     & 36.08     & 60.09     & 8.39      & 26.26     & 55.95 \\
    & 6             & 0.28      & 5.94      & 33.14     & 60.63     & 5.11      & 24.55     & 56.94 \\
    & \#G+\#O   & 0.32      & 4.54      & 31.24     & 59.78     & 2.87      & 21.35     & 56.68 \\
    \hline
    \multirow{4}{*}{JRDB-base~\cite{ehsanpour2022jrdb}}
    & 4             & 0.23     & 15.43      & 34.81     & 60.43     & 9.42      & 25.75     & 48.00 \\
    & 5             & 0.25     & 13.26      & 37.40     & 63.91     & 9.42      & 26.19     & 51.30 \\
    & 6             & 0.28     & 6.77       & 35.22     & 63.85     & 6.37      & 26.23     & 51.53 \\
    & \#G+\#O   & 0.32     & 4.49       & 34.40     & 61.46     & 3.15      & 25.80     & 49.71 \\
    \hline
    \multirow{4}{*}{HGC~\cite{tamura2022hunting}}        
    & 12            & 0.10     & 5.18       & 23.02     & 57.23     & 3.50      & 17.92     & 57.42 \\   
    & 24            & 0.10     & 5.60       & 21.44     & 54.57     & 3.00      & 14.48     & 53.64 \\   
    & 50            & 0.10     & 6.55       & 26.29     & 56.84     & 3.47      & 18.46     & 52.56 \\
    & 100           & 0.10     & 3.63       & 15.42     & 54.59     & 3.07      & 19.97     & 56.80 \\
    \hline
    \rowcolor{Gray}
    & 4             & 0.10     & 16.02              & \textbf{40.22}    & 64.06             
                               & 8.97               & 27.33             & 62.35 \\
    \rowcolor{Gray}
    & 8             & 0.10     & \underline{18.10}  & 37.51             & \underline{65.49} 
                               & \underline{9.79}   & \underline{29.23} & \textbf{63.93} \\
    \rowcolor{Gray}
    & 12            & 0.10     & \textbf{18.84}     & \underline{37.53} & \textbf{67.64}    
                               & \textbf{10.85}     & \textbf{30.90}    & \underline{63.84} \\
    \rowcolor{Gray}
    \multirow{-4}{*}{Ours}  
    & 16            &  0.10    & 15.03              & 37.03             & 65.31             
                               & 7.57               & 25.08             & 58.66 \\    
    \hline
\end{tabular}
}
\end{table}

\subsection{Comparison with the State of the Art}
\label{sec:5.2}
\noindent
\textbf{Compared methods.}
We compare our method with three clustering-based methods~\cite{wu2019learning,ehsanpour2020joint,ehsanpour2022jrdb} and one Transformer-based method~\cite{tamura2022hunting}. 
Since most GAD methods do not provide the official source code, we try our best to implement these previous work with necessary modifications for dealing with different problem settings, in particular the existence of outliers in Caf\'e.
Specifically, we adopt a fixed cluster size~\cite{ng2001spectral} for clustering-based methods since estimating the number of clusters did not perform well in Café due to the presence of spatially close outliers.
Note that no adjustment was made for datasets other than Caf\'e.

\noindent
$\bullet$ \textit{ARG}~\cite{wu2019learning}:
ARG utilizes graph convolutional 
networks~\cite{kipf2016semi} to model relations between actors in terms of position and appearance similarity. 
We apply spectral clustering~\cite{ng2001spectral} on a relation graph to divide actors into multiple groups. 

\noindent
$\bullet$ \textit{Joint}~\cite{ehsanpour2020joint} and \textit{JRDB-base}~\cite{ehsanpour2022jrdb}:
These models utilize GNNs to model relations between actors, and train actor representations to partition graphs by adopting a graph edge loss. 
JRDB-base further adopts geometric features.
Then, spectral clustering~\cite{ng2001spectral} is applied on the graph. 

\noindent
$\bullet$ \textit{HGC}~\cite{tamura2022hunting}:
Similar to our method, HGC employs a Transformer for GAD. 
However, unlike our method, HGC identifies group members by point matching between groups and actors on the 2D coordinate space. 
For a fair comparison, we utilize ground-truth actor tracklets to obtain actor features for HGC.

\begin{figure}[t]
\centering
\noindent
\begin{minipage}{1.0\linewidth}
\captionof{table}{Quantitative results on JRDB-Act validation-set.}
\label{table:jrdb-act}
\centering
\small
\scalebox{0.94}{
\begin{tabular}{lccccccc}
\hline
Method  & Backbone  & G1 AP     & G2 AP     & G3 AP     & G4 AP     & G5$^{+}$ AP   & mAP \\
\hline
SHGD~\cite{li2022self}            
        & Unipose~\cite{artacho2020unipose}         
                    & 3.1       & 25.0      & 17.5      & 45.6      & 25.2          & 23.3 \\
Joint~\cite{ehsanpour2020joint}           
        & I3D~\cite{carreira2017quo}       
                    & 8.0       & 29.3      & 37.5      & 65.4      & 67.0          & 41.4 \\
PAR~\cite{han2022panoramic}             
        & Inception-v3~\cite{szegedy2016rethinking}          
                    & 52.0      & 59.2      & 46.7      & 46.6      & 31.1          & 47.1 \\
JRDB-base~\cite{ehsanpour2022jrdb}       
        & I3D       & 81.4      & 64.8      & 49.1      & 63.2      & 37.2          & 59.2 \\
\rowcolor{Gray}
Ours            
        & ResNet-18~\cite{resnet} 
                    & 70.1      & 56.3      & 50.4      & 71.7      & 50.8          & \textbf{59.8} \\
\hline
\end{tabular}}
\end{minipage}
\begin{minipage}{.54\linewidth}
\captionof{table}{Quantitative results on Social-CAD.}
\label{table:social-cad}
\centering
\scalebox{0.8}{
\begin{tabular}{lccc}
\hline
Method                          & Backbone      & \# frames     & Social Accuracy   \\
\hline
ARG~\cite{wu2019learning}       & Inception-v3  & 17            & 49.0              \\
Joint~\cite{ehsanpour2020joint} & I3D           & 17            & 69.0              \\
\rowcolor{Gray}
Ours                            & ResNet-18     & 1             & \textbf{69.2}     \\
\hline
\end{tabular}}
\end{minipage}
\hspace{2mm}
\noindent
\begin{minipage}{.42\linewidth}
\captionof{table}{Ablation study on the group consistency loss.}
\label{table:abl_loss}
\centering
\scalebox{0.78}{
\setlength{\tabcolsep}{6pt}
\begin{tabular}{ccc}
\hline
$L_{\text{con}}$     & Group mAP$_{1.0}$        & Outlier mIoU   \\
\hline
\xmark                & 15.06                   & 63.35          \\
\rowcolor{Gray}
\cmark                & \textbf{18.84}          & \textbf{67.64} \\
\hline
\end{tabular}}
\end{minipage}
\begin{minipage}{.48\linewidth}
\captionof{table}{Ablation on the attention layers of the Grouping Transformer.}
\label{table:abl_attention}
\centering
\scalebox{0.74}{
\begin{tabular}{lcc}
\hline
Method                  & Group mAP$_{1.0}$     & Outlier mIoU      \\
\hline
Ours                    & \textbf{18.84}        & \textbf{67.64}    \\
w/o self-attention      & 13.53                 & 65.62    \\
w/o cross-attention     & 13.12                 & 64.19    \\
w/o grouping-attention  & 12.86                 & 64.65    \\
\hline
\end{tabular}
}
\end{minipage}
\hspace{2mm}
\noindent
\begin{minipage}{.48\linewidth}
\captionof{table}{Ablation on the use of distance mask and its threshold.}
\label{table:abl_mask}
\centering
\scalebox{0.74}{
\begin{tabular}{ccc}
\hline
Distance threshold ($\mu$)  & Group mAP$_{1.0}$ & Outlier mIoU   \\
\hline
0.1                         & 14.46             & 62.75          \\
\textbf{0.2}                & \textbf{18.84}    & \textbf{67.64}          \\
0.3                         & 15.09             & 63.49          \\
No threshold                & 14.96             & 67.03 \\
\hline
\end{tabular}}
\end{minipage}
\end{figure}

\noindent
\textbf{Caf\'e dataset.}
For a fair comparison, we use ImageNet pretrained ResNet-18 as the backbone and apply distance mask for all the methods including ours. 
We test every model on two different dataset splits as explained in \cref{sec:dataset_annotation}: 
\textit{split by view} and \textit{split by place}. 
Table~\ref{table:gad_performance} summarizes the results. 
Our model outperforms all the other methods by substantial margins on both splits in terms of both Group mAP and Outlier mIoU.
Note that the performance of clustering-based methods largely depends on the number of clusters, which is hard to determine or predict when there are outliers in a video clip. 
On the other hand, our model is less sensitive to the number of group tokens, 12 tokens shows the best performance on both settings though. 
Our model outperforms HGC, demonstrating the effectiveness of our group-actor matching in an embedding space, as opposed to the point matching strategy used in HGC. 
We also conduct experiments in a detection-based setting for all methods, and the results can be found in the appendix (\cref{sec:detection}).

\noindent
\textbf{JRDB-Act dataset.}
Table~\ref{table:jrdb-act} presents results on JRDB-Act.
Our model achieves $59.8$ mAP, surpassing all other methods. 
This indicates that our method effectively detects group activities across varying group sizes, especially when the group size is larger than 2.
Notably, our model with the ResNet-18 backbone outperforms Joint and JRDB-base with the substantially heavier I3D backbone.

\noindent
\textbf{Social-CAD dataset.}
Table~\ref{table:social-cad} summarizes the results on Social-CAD.
Our model surpasses the previous methods by using ResNet-18 with a single frame as backbone, which is significantly lighter than I3D backbone taking 17 frames as input in Joint model~\cite{ehsanpour2020joint}.
Due to the short length of video clips and small variations within clips, our model achieves the best even with a single frame.

\subsection{Ablation Studies}
We also verify the effectiveness of our proposed model through ablation studies on Caf\'e, \textit{split by view} setting.

\noindent
\textbf{Impact of the proposed loss function.}
Table~\ref{table:abl_loss} shows the effectiveness of the group consistency loss, 
which improves Group mAP by a substantial margin. 
This result demonstrates that the group consistency loss, which brings actor embeddings within the same group closer, has a significant impact for GAD.
We do not ablate the other losses since they are inevitable for the training.

\noindent
\textbf{Effects of the attention layers.}
Table~\ref{table:abl_attention} summarizes the effects of multi-head attention layers in Grouping Transformer. 
Note that self-attention in this table stands for both multi-head self-attention layers that captures relationship between actors and those between groups, cross-attention means multi-head cross-attention layers that actor features attends frame-level features to capture contextual information, and grouping-attention refers to the layers that group tokens attends actor features to form group representation. 
The results demonstrate that all three attention layers contribute to the performance.
Particularly, removing grouping-attention layer results in the largest performance drop in Group mAP since grouping-attention layer learns the relationship between group embeddings and actor embeddings, aiding in group localization.

\noindent
\textbf{Effects of the distance mask.}
We investigate the efficacy of utilizing the distance mask. 
Distance mask inhibits self-attention between a pair of actors whose distance is greater than the distance threshold $\mu$. 
As shown in Table~\ref{table:abl_mask}, applying distance mask between actors is effective in most cases but too small threshold, 0.1 in this table, degrades the performance. 
It is because actors can interact only with nearby actors at small distance threshold, which might mask the interaction between actors of the same group. 
Distance threshold of 0.2 reaches the best result while slightly degrades at 0.3.

\begin{figure}[t!]
\centering
\includegraphics[width=1.0\linewidth]{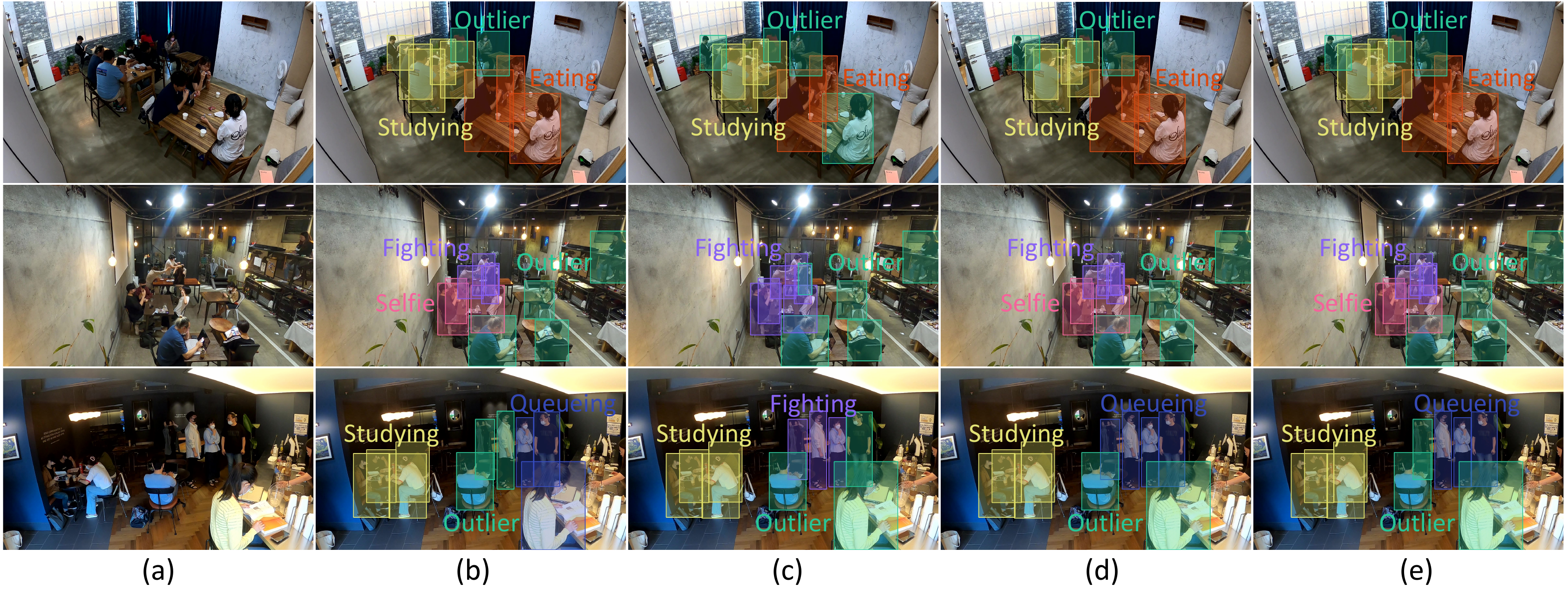}
\caption{Qualitative results on Caf\'e test-set, \textit{split by view} setting. 
Boxes with the same color belong to the same group. 
(a) Input frame. 
(b) Prediction of JRDB-base. 
(c) Prediction of HGC. 
(d) Prediction of our model.
(e) Ground-truth.
}
\label{fig:qual}
\end{figure}

\subsection{Qualitative Analysis}
\cref{fig:qual} visualizes the predictions of JRDB-base, HGC, and our model.
The results show that our model is able to localize multiple groups and predict their activity classes at the same time, and more reliably than the others, even in challenging 
densely populated scenes with a lot of outliers.

%% file: arxiv_sections/6_conclusion.tex
\section{Conclusion}
We have introduced a new challenging benchmark, dubbed Caf\'e, and a new model based on Transformer to present a direction towards more practical GAD.
As Caf\'e exhibits multiple non-singleton groups per clip and provides rich annotations of actor bounding boxes, track IDs, group configurations, and group activity labels, it can serve as a new, practical, and challenging benchmark for GAD. 
Also, the proposed model can deal with a varying number of groups as well as predicting members of each group and its activity class. 
Our model outperformed prior arts on three benchmarks including Caf\'e.
We believe that our dataset and model will promote future research on more practical GAD.

\noindent \textbf{Limitation:} Our model does not consider much about temporal and multi-view aspects of the proposed dataset. Improving upon these aspects will be a valuable direction to explore.

%% file: arxiv_supp_sections/0_introduction.tex
This appendix presents contents omitted in the main paper due to the page limit. 
Sec.~\ref{sec:dataset} presents more details of our proposed Caf\'e dataset. 
In Sec.~\ref{sec:comparison}, we provide a detailed explanation of the statistics used for comparison with other group activity detection datasets in \cref{sec:dataset_comparison} of the main paper.
Further analysis and more experiments are given in Sec.~\ref{sec:more_experiments}.

%% file: arxiv_supp_sections/1_dataset.tex
\section{More Detailed Information of Caf\'e}
\label{sec:dataset}

\subsection{Data Acquisition}
To construct practical group activity detection (GAD) dataset, we recruit participants, direct their actions, and record using cameras, with the assistance from Deeping Source (\url{https://www.deepingsource.io/}). 
We take videos at six different cafes, installing cameras at four different locations within each cafe to capture the same scene from four distinct viewpoints. 
Videos of Caf\'e are recorded with RGB cameras at $1920 \times 1080$ resolution, then are cut into clips at 6-seconds intervals. 
Members of each group and clothes of participants change periodically to prevent monotonous grouping patterns.
Note that we obtain informed consent of participants to release the dataset publicly.

\begin{figure}[p] 
\begin{center}
\includegraphics[width=\textwidth]{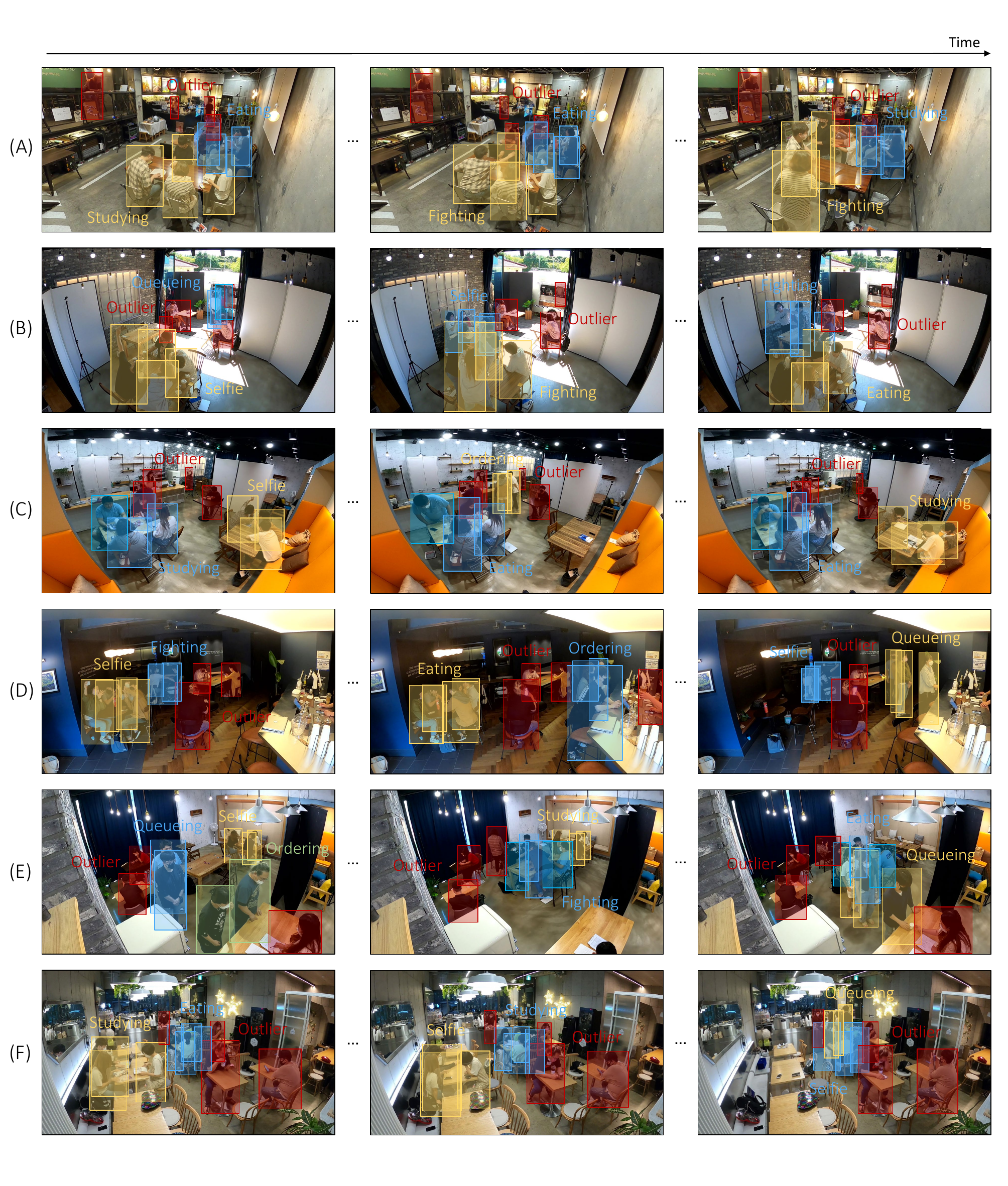}    
\end{center}
\caption{Qualitative examples of videos from Caf\'e for each place along with annotations. 
Different color represents different groups, with all outliers are marked in red.}
\label{fig:annotation}
\end{figure}

In each video clip, actors are divided into multiple groups and outliers. 
Members of each group are asked to perform one of predefined group activities: 
\textit{queueing}, \textit{ordering}, \textit{eating}, \textit{working}, \textit{fighting}, and \textit{taking selfie}.
On the other hand, outliers who do not belong to any group, perform arbitrary action (\eg~look at a smart phone or work on a laptop).
Note that actors who are close together but do not interact with any group members are also considered as outliers. 
Therefore, localizing group members necessitates understanding the properties of individual actors and their relationship as well as their spatial proximity.

Fig.~\ref{fig:annotation} provides qualitative examples of videos from Caf\'e, showing each location along with annotations that include bounding boxes of humans, group configurations, and group activity labels.
Each place of Caf\'e exhibits different characteristics such as varying camera viewpoints, group configurations, and light conditions. 
Notably, one of six places in Caf\'e includes outdoor scenes (Fig.~\ref{fig:annotation} (B)), which diversify the scene variety of Caf\'e. 
All video frames and annotations is available at \url{https://cvlab.postech.ac.kr/research/CAFE}.

\setlength{\tabcolsep}{8pt}
\begin{table}[t!]
\begin{center}
\caption{Statistics of Caf\'e \textit{split by place} training-validation-test split. \# represents ``the number of''. Example videos and annotations of each place are illustrated in Fig.~\ref{fig:annotation}. }
\label{tab:dataset_place}
\begin{tabular}{cccccc}
\hline
Place           & \# Clips      & \# Frames     & \# Boxes      & \# Groups     \\
\hline
\multicolumn{5}{c}{\textbf{Training}}\\
\hline
A               & 1,851          & 59,246         & 620,194        & 2,628          \\
B               & 1,628          & 55,603         & 555,517        & 2,455          \\
C               & 1,616          & 55,369         & 601,309        & 3,117          \\
D               & 1,614          & 55,194         & 576,952        & 2,630          \\
\hline
\multicolumn{5}{c}{\textbf{Validation}}\\
\hline
E               & 1,423          & 48,291         & 510,868        & 2,224          \\
\hline
\multicolumn{5}{c}{\textbf{Test}}\\
\hline
F               & 2,165          & 69,756         & 742,458        & 2,917          \\
\hline
\end{tabular}
\end{center}
\end{table}

\setlength{\tabcolsep}{8pt}
\begin{table}[t!]
\begin{center}
\caption{Statistics of Caf\'e \textit{split by view} training-validation-test split. \# represents ``the number of''. There are 4 different camera viewpoints for each place, it is named $\alpha$, $\beta$, $\gamma$, and $\delta$, respectively.}
\label{tab:dataset_view}
\begin{tabular}{cccccc}
\hline
View            & \# Clips      & \# Frames     & \# Boxes      & \# Groups     \\
\hline
\multicolumn{5}{c}{\textbf{Training}}\\
\hline
$\alpha$          & 2,574          & 85,901         & 898,486        & 3,940          \\
$\beta$           & 2,572          & 85,777         & 916,590        & 4,056          \\
\hline
\multicolumn{5}{c}{\textbf{Validation}}\\
\hline
$\gamma$          & 2,576          & 86,049         & 920,179        & 4,063          \\
\hline
\multicolumn{5}{c}{\textbf{Test}}\\
\hline
$\delta$          & 2,575          & 85,732         & 872,043        & 3,912          \\
\hline
\end{tabular}
\end{center}
\end{table}

\subsection{Additional Dataset Statistics}
\label{sec:a.2}
Additional statistics of \textit{split by place} and \textit{split by view} training-validation-test split are shown in Table~\ref{tab:dataset_place} and Table~\ref{tab:dataset_view}, respectively. 
For the \textit{split by place} setting, 24 videos are split into 16 training videos of 4 places, 4 validation videos of 1 place, and 4 test videos of 1 place.
On the other hand, for the \textit{split by view} setting, 4 videos of each place are split into 2 training, 1 validation, and 1 test videos. 
In a total of 6 places, 24 videos are split into 12 training, 6 validation, and 6 test videos.
The number of clips, frames, boxes, and groups are evenly distributed across each place and camera viewpoint. 
For the entire dataset,
the average number of frames per clip is 33.36, the average number of boxes per frame is 10.5, and the average number of groups per clip is 1.55. 
The original videos have an average FPS of 29.9.

\subsection{Rationale for Caf\'e Scenes and Group Activity Classes}
Cafes serve as appropriate environments for simulating realistic daily group activities, since people tend to gather in groups and multiple groups engage in their own activities simultaneously. 
We take videos at a total of six cafes (Fig.~\ref{fig:annotation}), ensuring scene variety by including one cafe with an outdoor scene (Place B) and another with very low light condition (Place D).

Concerning group activity classes, we define them as activities performed collectively by a group of people. 
Thus, we also include classes such as \textit{eating} and \textit{working}, which can also be done individually. 
While solely using indoor scenes may limit diversity, it is worth noting that even daily outdoor videos exhibit a constrained range of group activity classes.
Moreover, existing datasets~\cite{choi2009they,ibrahim2016hierarchical,yan2020social,ehsanpour2020joint,ehsanpour2022jrdb} have relatively fewer group activity classes compared to individual action classes. 
For instance, JRDB-Act dataset~\cite{ehsanpour2022jrdb} includes only three person-to-person activity classes among a total of 26 action classes.

\section{Statistics used for Dataset Comparison}
\label{sec:comparison}
In this section, we will explain the detailed explanation of the statistics used for comparison with other GAD datasets in \cref{sec:dataset_comparison} of the main paper, which are aspect ratio ({Fig.}~3b), population density ({Fig.}~3c), and inter-group distance ({Fig.}~3d).

\subsubsection{Aspect Ratio.}
The aspect ratio or height-to-width ratio is defined as the ratio of raw pixel height to the raw pixel width of each actor bounding box. 

\subsubsection{Population Density.}
Population density is calculated by first creating a convex hull that encloses the bounding boxes of actors participating in group activities within each video clip.
Next, we define population density for each clip as the ratio of the union area occupied by the actors participating in group activities to the area of the convex hull.
Finally, the population densities are averaged over all clips.
Higher population density poses challenges for both group localization and classification, due to the proximity or occlusion of other groups or outliers.

\subsubsection{Inter-group Distance.}
\label{sec:inter_distance}
We measure inter-group distance by calculating the distance between each group and its nearest group or outlier as follows:

\begin{equation}
    \text{Inter-group distance} = \frac{1}{|G|} \sum_{g \in G} \min_{i \in g, j \notin g}\frac{{\lVert c_i - c_j \rVert}_2}{\sqrt{\frac{1}{2}(\text{Area}_i + \text{Area}_j)}}, 
\end{equation}
where $|G|$ is the set of groups for all datasets, $c_i$ and $c_j$ denote the center coordinates of boxes $i$ and $j$, $\text{Area}_i$ and $\text{Area}_j$ represent the areas of boxes $i$ and $j$. 
An actor $i$ is a member of group $g$, while an actor $j$ appears in the same clip as $g$ but is not a member of $g$.
To account for viewpoint variations, we normalize the $\ell_2$ distance using the average area of the two bounding boxes. 
Finally, theses distances are averaged across all groups to obtain the inter-group distance for each dataset. 
As mentioned in \cref{sec:dataset_comparison} of the main paper, a small inter-group distance of a dataset implies the presence of other actors nearby who are not part of the group, indicating the importance for understanding semantic relationships alongside spatial proximity in group localization.

%% file: arxiv_supp_sections/2_experiments.tex
\section{More Experimental Results}
\label{sec:more_experiments}

\begin{figure}[t!]
    \centering    
    \subfloat[\centering \label{fig:confusion_view}]{{\includegraphics[width=0.48\linewidth]{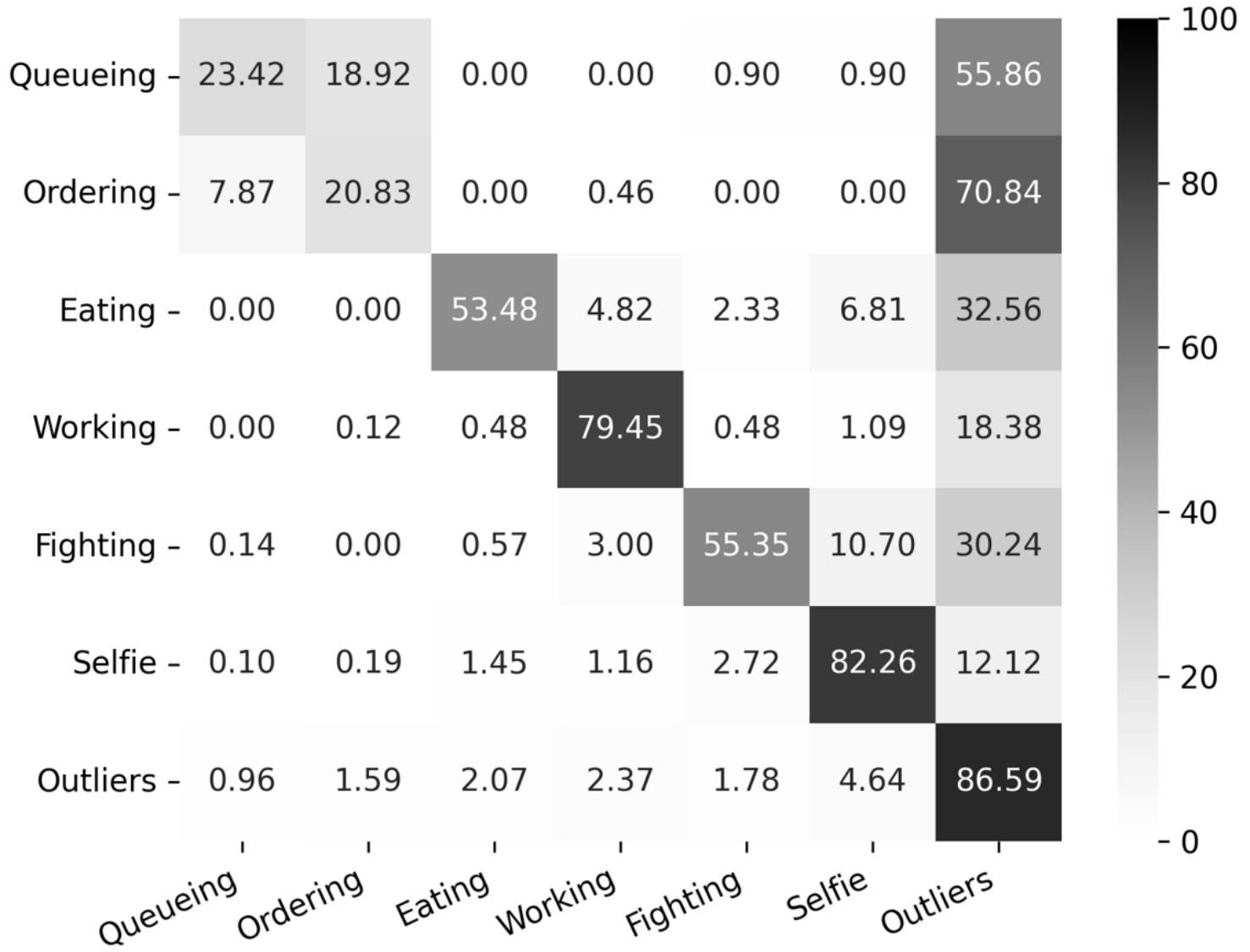}}}    
    \hfill    
    \subfloat[\centering \label{fig:confusion_place}]{{\includegraphics[width=0.48\linewidth]{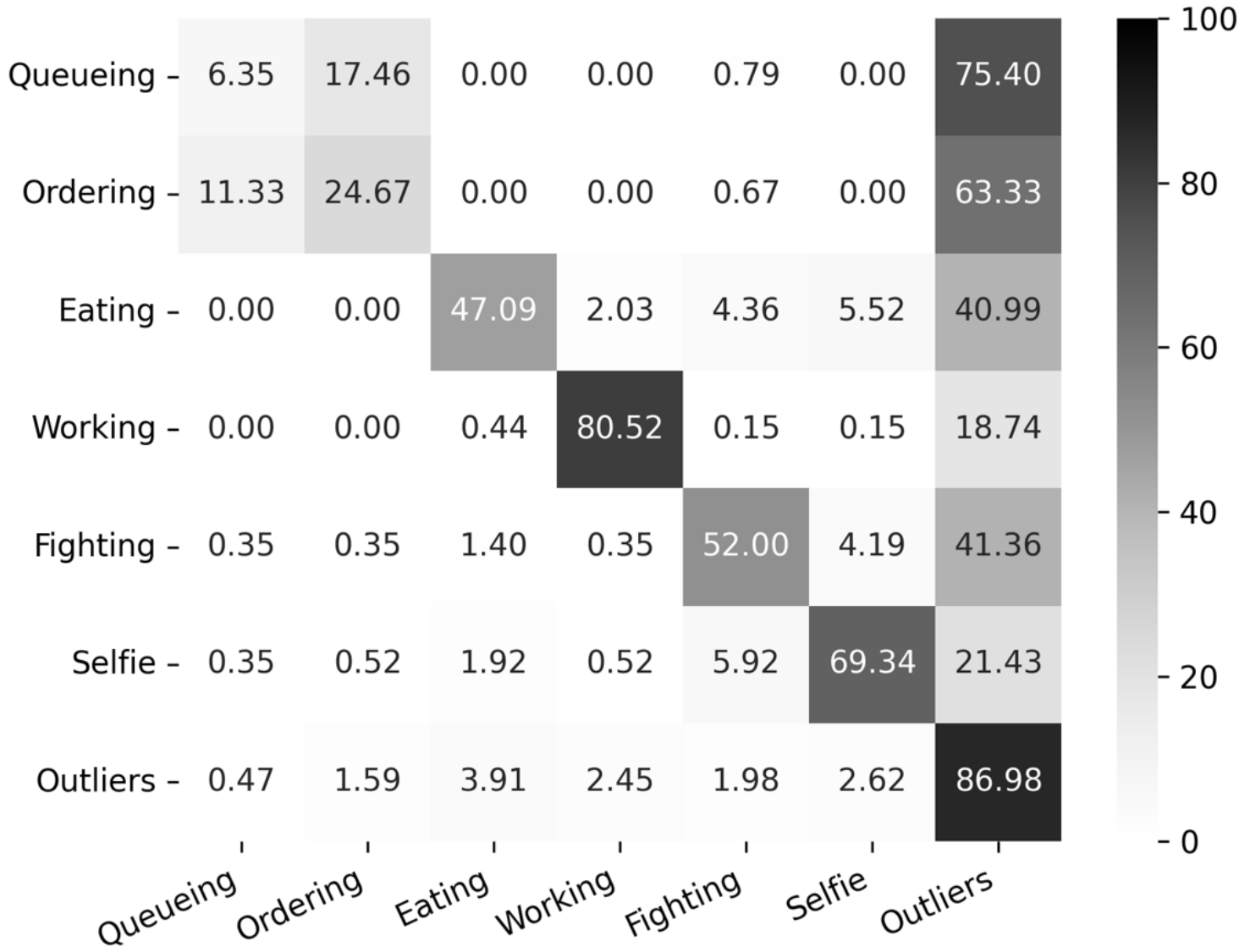}}}
    \caption{The confusion matrix on Caf\'e: 
    (a) \textit{split by view} setting, 
    (b) \textit{split by place} setting. 
    }
    \label{fig:confusion}
\end{figure}

\subsection{Analysis of Activity Classification}
\label{sec:analysis}
Fig.~\ref{fig:confusion} shows the confusion matrix for two distinct dataset splits of Caf\'e: \textit{split by view} and \textit{split by place}. 
These confusion matrices are computed based on predicted group and ground-truth group pairs with Group IoU greater than or equal to 0.5. 
Notably, our method shows weak performance on \textit{queueing} and \textit{ordering} classes due to the class imbalance issue; theses classes are the least frequent and the second least frequent group activity class in Caf\'e, respectively. 
Consequently, a significant portion of these groups is misclassified as outliers. 
Addressing this challenge in the future could involve weighting the group activity classifier with the inverse of class frequency or adopting focal loss~\cite{lin2017focal}.
Across both dataset splits, the most confusing group activity classes to distinguish are \textit{queueing} versus \textit{ordering}, as they frequently co-occur.

\subsection{Detection-based Setting on Caf\'e}
\label{sec:detection}
To show the effectiveness of our proposed method in realistic situation, we replace the ground-truth actor tracklets with the predicted actor tracklets which are obtained by utilizing off-the-shelf multi-object tracker~\cite{zhang2021bytetrack}.

\subsubsection{Tracklet Matching.}
Similar to group matching explained in \cref{sec:4.2} of the main paper, we need another bipartite matching between ground-truth tracklets and predicted tracklets to train the model in detection-based setting. 
Among all possible combinations of $N$ ground-truth tracklets and $M$ predicted tracklets ($M \geq N$ in our case), Hungarian algorithm finds the permutation with the lowest matching cost:
\begin{equation}
    \hat{\sigma} = \underset{\sigma \in \mathfrak{S}_M}{\arg\min} \sum_{i}^{N} C_{i, \sigma(i)}, \label{eq:hungarian_tracklet}
\end{equation}
where the matching cost of the ground-truth tracklet $i$ and the predicted tracklet $\sigma(i)$ is defined as
\begin{equation}
    C_{i, \sigma(i)} = \sum_{t=1}^{T} \big(\lambda_\text{L1} \|b_{i, t} - \hat{b}_{\sigma(i), t}\|_{1} + \lambda_\text{giou}L_\text{giou}(b_{i, t}, \hat{b}_{\sigma(i), t}) \big),
\end{equation}
\noindent
where $b_{i,t}$ is the bounding box of ground-truth tracklet $i$ at frame $t$, $\hat{b}_{\sigma(i), t}$ is the bounding box of predicted tracklet $\sigma(i)$ at frame $t$, $T$ is the number of frames for each video clip, $L_\text{giou}$ is the generalized IoU loss~\cite{rezatofighi2019generalized}, $\lambda_{L1}=5.0$, and $\lambda_{giou}=2.0$.

\subsubsection{Evaluation.}
To measure the performance in the detection-based setting, we need identity matching between the predicted tracklets and the ground-truth tracklets. 
We follow standard box matching strategy used in object detection, utilize the mean IoU over $T$ frames as the IoU threshold and the mean confidence over $T$ frames as the detection confidence score of the predicted tracklet. 
We adopt IoU threshold of 0.5 following previous methods~\cite{ehsanpour2020joint,ehsanpour2022jrdb}.

\setlength{\tabcolsep}{7pt}
\begin{table}[t!]
\caption{Comparison with the previous GAR and GAD models on Caf\'e detection-based setting. We mark the best and the second-bset performance in \textbf{bold} and \underline{underline}, respectively.}
\label{table:det_setting}
\centering
\small
\begin{tabular}{lccccccc}
    \hline
    \multirow{3}{*}{Method} 
    & \multicolumn{3}{c}{Split by view} & \multicolumn{3}{c}{Split by place}\\
    \cline{2-7}    
    & \makecell[c]{Group \\ mAP$_{1.0}$}    & \makecell[c]{Group \\ mAP$_{0.5}$}    & \makecell[c]{Outlier \\ mIoU}     
    & \makecell[c]{Group \\ mAP$_{1.0}$}    & \makecell[c]{Group \\ mAP$_{0.5}$}    & \makecell[c]{Outlier \\ mIoU}     \\
    \hline
    ARG~\cite{wu2019learning}         
                & 7.92                      & 29.83                     & 48.82             
                & 6.70                      & \underline{26.27}         & 48.32     \\            
    Joint~\cite{ehsanpour2020joint}       
                & 9.14                      & 31.83                     & 42.93             
                & 6.08                      & 18.43                     & 2.83      \\    
    JRDB-base~\cite{ehsanpour2022jrdb}   
                & \underline{12.63}        & \underline{35.53}          & 31.85             
                & \underline{8.15}          & 22.68                     & 33.03     \\    
    HGC~\cite{tamura2022hunting}         
                & 6.77                      & 31.08                     & \underline{57.65}
                & 4.27                      & 24.97                     & \underline{57.70}     \\
    \rowcolor{Gray}
    Ours        & \textbf{14.36}            & \textbf{37.52}            & \textbf{63.70}          
                & \textbf{8.29}             & \textbf{28.72}            & \textbf{59.60} \\
    \hline
\end{tabular}
\end{table}

\subsubsection{Comparison with the Previous Work.}
We implement previous GAR and GAD methods~\cite{ehsanpour2020joint, tamura2022hunting, wu2019learning} as explained in \cref{sec:5.2} of the main paper and test our proposed model and other methods on two different dataset splits: \textit{split by view} and \textit{split by place}. 
We adopt three evaluation metrics: Group mAP$_{1.0}$, Group mAP$_{0.5}$, and Outlier mIoU. 
Group mAP$_{1.0}$ is the primary measure for GAD, a predicted group is considered as true positive only when including correct members only. 
Group mAP$_{0.5}$ allows small mistakes on group localization, and Outlier mIoU measures outlier detection performance. 
Table~\ref{table:det_setting} summarizes the results of detection-based setting on Caf\'e. 
Our model outperforms other methods on both dataset splits. 
Although JRDB-base~\cite{ehsanpour2022jrdb} is comparable to our model on \textit{split by place} setting in terms of Group mAP$_{1.0}$, our model largely outperforms in terms of Group mAP$_{0.5}$ and Outlier mIoU.
While the performance of most methods including ours drop in detection-based setting when compared to the use of ground-truth actor tracklets, the performance of HGC does not drop as it employs box prediction loss which can be used to refine actor features obtained from incorrect predicted actor tracklets. 
Nevertheless, our proposed method still outperforms HGC even when evaluated in detection-based setting.

\setlength{\tabcolsep}{3pt}
\begin{table}[t!]
\caption{Comparison on Caf\'e using previous metrics.}
\small
\centering
\scalebox{0.85}{
\begin{tabular}{ccccccccc}
\hline
\multirow{2}{*}{Method} & \multicolumn{4}{c}{Split by view} & \multicolumn{4}{c}{Split by place}\\
\cline{2-9} 
    & \footnotesize{G mAP$_{1.0}$}  
    & \footnotesize{Social acc} & \footnotesize{F$_1$} & \footnotesize{Indiv acc}     
    & \footnotesize{G mAP$_{1.0}$}  
    & \footnotesize{Social acc} & \footnotesize{F$_1$} & \footnotesize{Indiv acc}  \\
\hline
ARG~\cite{wu2019learning}         
            & 11.03 & 67.85 & 65.80 & 69.89 
            & 6.87  & 60.23 & 58.01 & 61.66 \\
Joint~\cite{ehsanpour2020joint}       
            & 14.05 & 64.98 & 56.91 & 68.43
            & 8.39  & 61.14 & 53.88 & 63.50 \\
JRDB-base~\cite{ehsanpour2022jrdb}   
            & 15.43 & 67.58 & 64.19 & 68.56
            & 9.42  & 57.59 & 53.65 & 58.36 \\
HGC~\cite{tamura2022hunting}
            & 6.55  & 61.88 & 53.21 & 62.87
            & 3.47  & 57.95 & 41.04 & 58.85 \\
\rowcolor{Gray}
Ours        & \textbf{18.84}    & \textbf{71.51}    
            & \textbf{67.55}    & \textbf{72.65}    
            & \textbf{10.85}    & \textbf{68.43}    
            & \textbf{62.58}    & \textbf{69.02}    \\
\hline
\end{tabular}}
\label{tab:other_metrics}
\end{table}

\setlength{\tabcolsep}{7pt}
\begin{table}[t!]
\caption{Comparison on Caf\'e treating outliers as singletons.}
\centering
\scalebox{0.85}{
\begin{tabular}{ccccc}
\hline
\multirow{2}{*}{Method} 
    & \multicolumn{2}{c}{Split by view} & \multicolumn{2}{c}{Split by place}\\
    \cline{2-5}    
    & Group mAP$_{1.0}$    & Group mAP$_{0.5}$
    & Group mAP$_{1.0}$    & Group mAP$_{0.5}$ \\
\hline
ARG~\cite{wu2019learning}           & 9.36  & 29.79 & 5.81  & \underline{24.47}  \\
Joint~\cite{ehsanpour2020joint}     & 11.74 & \underline{32.07} & 7.14  & 23.58  \\
JRDB-base~\cite{ehsanpour2022jrdb}  & \underline{13.23} & 30.39 & \underline{8.03}  & 22.24  \\
HGC~\cite{tamura2022hunting}        & 8.33  & 24.27 & 6.72  & 18.88  \\
\rowcolor{Gray}
Ours        & \textbf{18.80}    & \textbf{36.27}    & \textbf{11.14}    & \textbf{29.15} \\
\hline
\end{tabular}}
\label{tab:including_outlier}
\end{table}

\subsection{Comparison with additional evaluation metrics}
\label{sec:more_metrics}

To further demonstrate the effectiveness of the proposed method, we conduct additional comparisons using metrics employed in previous work, such as social accuracy~\cite{ehsanpour2020joint}, $\text{F}_{1}$ score~\cite{han2022panoramic}, and individual accuracy. 
Table~\ref{tab:other_metrics} summarizes the results. 
Our model surpasses the existing models in all measures of both dataset split by a substantial margin. 

In Table~\ref{tab:including_outlier}, we present Group mAP scores considering outliers as singleton groups. 
The results indicate that our model outperforms other models even when outliers are considered as singleton groups.
Specifically, ours and HGC~\cite{tamura2022hunting} demonstrate similar or slightly improved performance when including outliers as groups, while clustering-based methods~\cite{wu2019learning,ehsanpour2020joint,ehsanpour2022jrdb} show a decrease in performance. 
This is likely due to the difficulty of accurately separating spatially-close outliers using off-the-shelf clustering.

\subsection{More Qualitative Results}
\label{sec:more_qual}
\begin{figure}[p] 
\centering
\includegraphics[width=\textwidth]{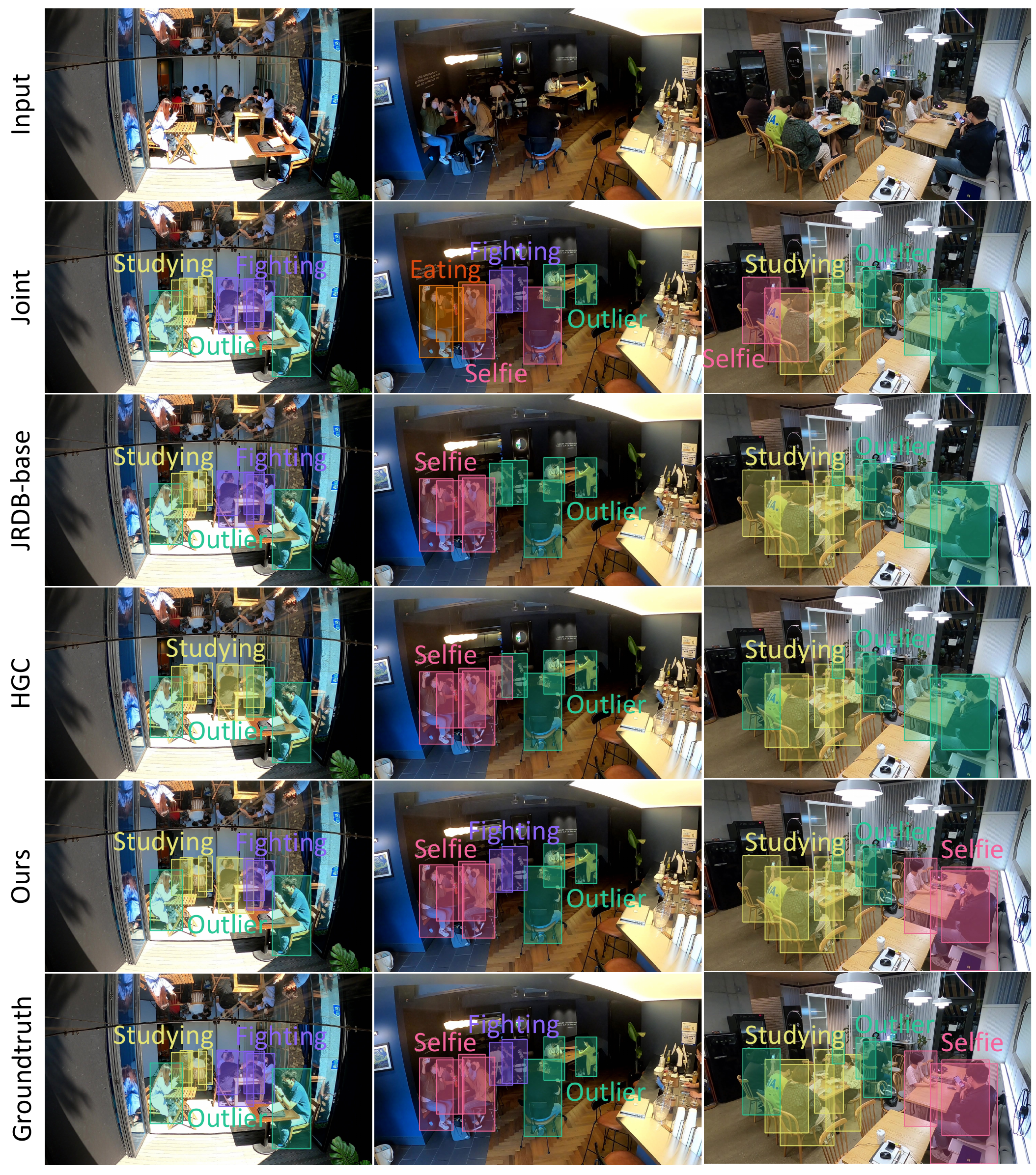}    
\caption{More qualitative results on Caf\'e test-set, \textit{split by view} setting. Boxes with the same color belong to the same group.}
\label{fig:qualitative}
\end{figure}

Fig.~\ref{fig:qualitative} visualizes the predictions of Joint~\cite{ehsanpour2020joint}, JRDB-base~\cite{ehsanpour2022jrdb}, HGC~\cite{tamura2022hunting}, and our model. 
The results show that our model is able to localize multiple groups more reliably than others, such as \textit{fighting} group for second example and \textit{taking selfie} group for third example.

